\RequirePackage{fix-cm}
\documentclass[english,twocolumn]{svjour3}
\usepackage[T1]{fontenc}
\usepackage[latin9]{inputenc}
\usepackage{color}
\usepackage{babel}
\usepackage{array}
\usepackage{refstyle}
\usepackage{booktabs}
\usepackage{units}
\usepackage{textcomp}
\usepackage{url}
\usepackage{multirow}
\usepackage{amsmath}
\usepackage{amssymb}
\usepackage{graphicx}
\usepackage{microtype}
\usepackage[unicode=true,pdfusetitle,
 bookmarks=true,bookmarksnumbered=false,bookmarksopen=false,
 breaklinks=false,pdfborder={0 0 1},backref=false,colorlinks=true]
 {hyperref}
\usepackage{biblatex-files}

\makeatletter

\AtBeginDocument{\providecommand\figref[1]{\ref{fig:#1}}}
\AtBeginDocument{\providecommand\secref[1]{\ref{sec:#1}}}
\AtBeginDocument{\providecommand\subsecref[1]{\ref{subsec:#1}}}
\AtBeginDocument{\providecommand\Eqref[1]{\ref{Eq:#1}}}
\AtBeginDocument{\providecommand\tabref[1]{\ref{tab:#1}}}
\providecommand{\tabularnewline}{\\}
\RS@ifundefined{subsecref}
  {\newref{subsec}{name = \RSsectxt}}
  {}
\RS@ifundefined{thmref}
  {\def\RSthmtxt{theorem~}\newref{thm}{name = \RSthmtxt}}
  {}
\RS@ifundefined{lemref}
  {\def\RSlemtxt{lemma~}\newref{lem}{name = \RSlemtxt}}
  {}

\usepackage{graphicx}
\smartqed

\usepackage{csquotes}

\newref{sec}{%
name = \RSsectxt,
names = \RSsecstxt,
Name = \RSSectxt,
Names = \RSSecstxt,
refcmd = {\ref{#1}},  
rngtxt = \RSrngtxt,
lsttwotxt = \RSlsttwotxt,
lsttxt = \RSlsttxt}

\newref{subsec}{%
name = \RSsectxt,
names = \RSsecstxt,
Name = \RSSectxt,
Names = \RSSecstxt,
refcmd = {\ref{#1}},  
rngtxt = \RSrngtxt,
lsttwotxt = \RSlsttwotxt,
lsttxt = \RSlsttxt}

\newref{appendix}{%
name = \RSappendixname,
names = \RSappendicesname,
Name = \RSAppendixname,
Names = \RSAppendicesname,
refcmd = {\ref{#1}},  
rngtxt = \RSrngtxt,
lsttwotxt = \RSlsttwotxt,
lsttxt = \RSlsttxt}

\usepackage{algpseudocode}
\algtext*{EndWhile}
\algtext*{EndIf}
\algtext*{EndFor}
\algtext*{EndProcedure}

\@ifundefined{showcaptionsetup}{}{%
 \PassOptionsToPackage{caption=false}{subfig}}
\usepackage{subfig}
\makeatother

\usepackage[style=apa]{biblatex}

\begin{document}
\title{Provident Vehicle Detection at Night for Advanced Driver Assistance Systems}


\institute{       
L.\ Ewecker \at                
Dr.\ Ing.\ h.c.\ F.\ Porsche AG, Weissach, Germany.
\email{lukas.ewecker@porsche.de}
\and
E.\ Asan$^*$ \at
Robert Bosch GmbH, Leonberg, Germany.
\email{ebubekir.asan@bosch.com}
\and
L.\ Ohnemus$^*$ \at 
Karlsruhe Institute of Technology, Karlsruhe, Germany.
\email{lars.ohnemus@student.kit.edu}
\and
S.\ Saralajew$^*$ \at               
NEC Laboratories Europe GmbH, Heidelberg, and Leibniz University Hannover, Institute of Product Development, Hannover, both Germany. 
\email{sascha.saralajew@neclab.eu}
\and
$^*$\,The research was performed during employment at Dr.\ Ing.\ h.c.\ F.\ Porsche AG.\\
\\
\emph{Authors contributed equally.}
}
\author{Lukas Ewecker \and Ebubekir Asan \and Lars Ohnemus \and Sascha
Saralajew}

\maketitle
\begin{abstract}
In recent years, computer vision algorithms have become more powerful,
which enabled technologies such as autonomous driving to evolve rapidly.
However, current algorithms mainly share one limitation: They rely
on directly visible objects. This is a significant drawback compared
to human behavior, where visual cues caused by objects (e.\,g., shadows)
are already used intuitively to retrieve information or anticipate
occurring objects. While driving at night, this performance deficit
becomes even more obvious: Humans already process the light artifacts
caused by the headlamps of oncoming vehicles to estimate where they
appear, whereas current object detection systems require that the
oncoming vehicle is directly visible before it can be detected. Based
on previous work on this subject, in this paper, we present a complete
system that can detect light artifacts caused by the headlights of
oncoming vehicles so that it detects that a vehicle is approaching
providently (denoted as provident vehicle detection). For that, an
entire algorithm architecture is investigated, including the detection
in the image space, the three-dimensional localization, and the tracking
of light artifacts. To demonstrate the usefulness of such an algorithm,
the proposed algorithm is deployed in a test vehicle to use the detected
light artifacts to control the glare-free high beam system proactively
(react before the oncoming vehicle is directly visible). Using this
experimental setting, the provident vehicle detection system's time
benefit compared to an in-production computer vision system is quantified.
Additionally, the glare-free high beam use case provides a real-time
and real-world visualization interface of the detection results by
considering the adaptive headlamps as projectors. With this investigation
of provident vehicle detection, we want to put awareness on the unconventional
sensing task of detecting objects providently (detection based on
observable visual cues the objects cause before they are visible)
and further close the performance gap between human behavior and computer
vision algorithms to bring autonomous and automated driving a step
forward.

\end{abstract}

\section{Introduction\label{sec:Introduction}}

Humans have five senses, and, out of those, visual perception is likely
to be the primary information relevant for driving a vehicle \parencite{Sivak1996}.
This fact is one reason why visual perception is essential in robots
and assisted (autonomous) driving. In the last years, visual perception
by machines has made tremendous progress by using Neural Networks
(NNs; e.\,g., \cite{fasterrcnn,yolo3,imagenet}) and achieved superhuman
performance for some tasks \parencite{resnet}. However, while progressing
heavily in specific directions like object detection or action recognition,
the abilities are far behind the general-purpose human performance,
which is, for instance, reflected in the problem of adversarial robustness
\parencite[e.\,g.,][]{Eykholt2018}.

\begin{figure*}
\includegraphics[scale=0.23]{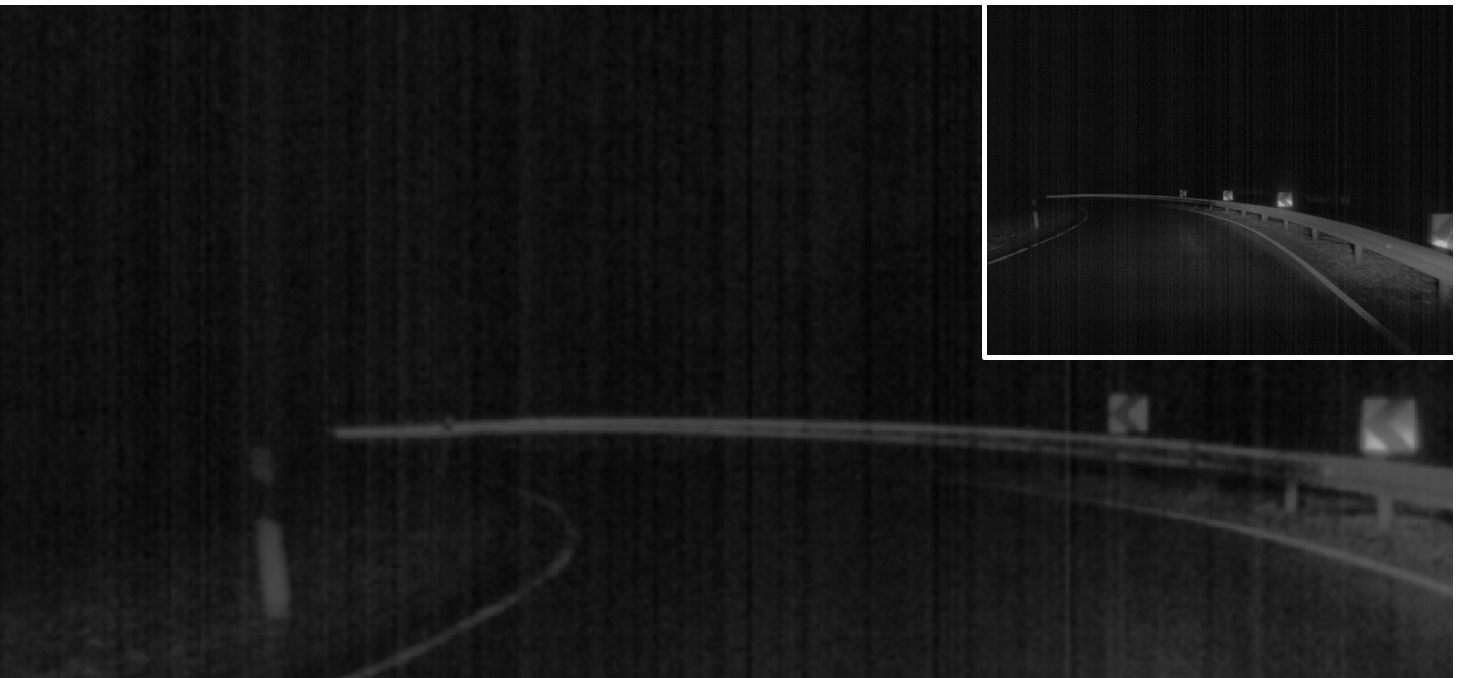}\hfill{}\includegraphics[scale=0.23]{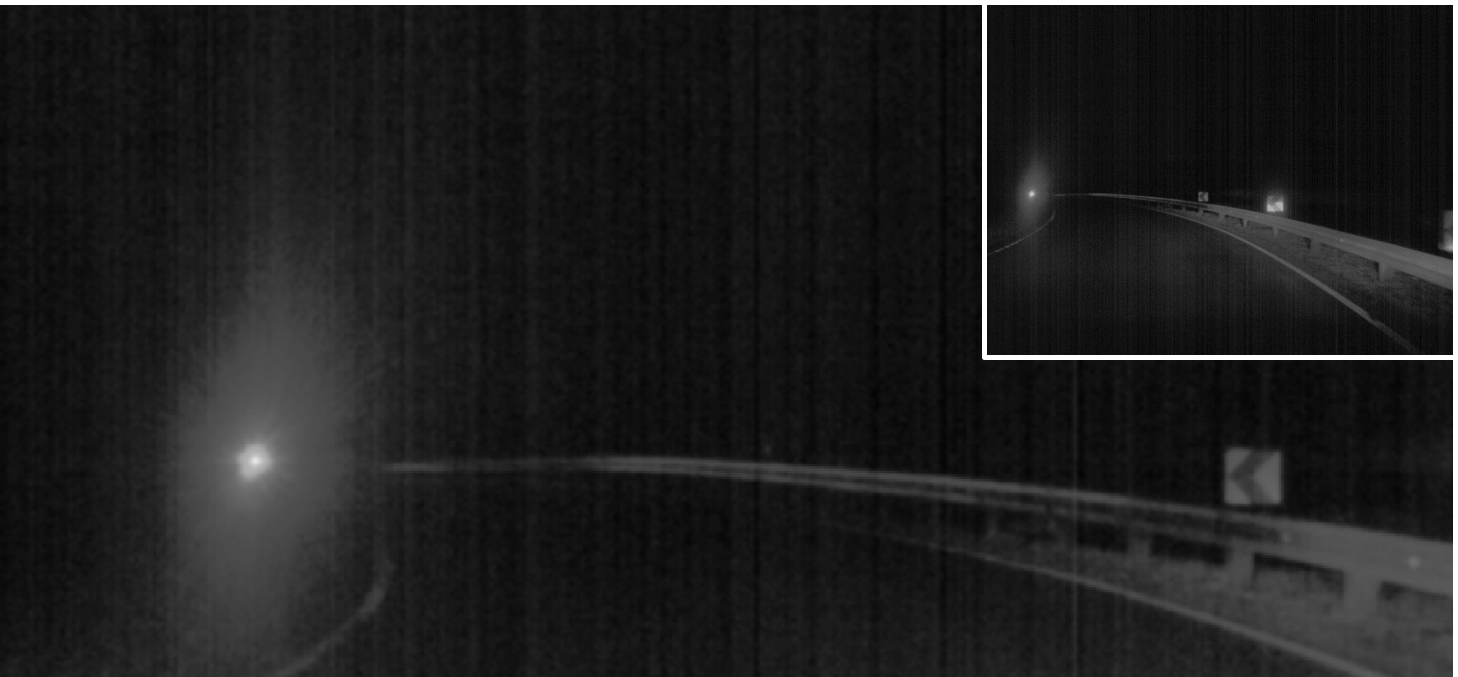}\hfill{}\includegraphics[scale=0.23]{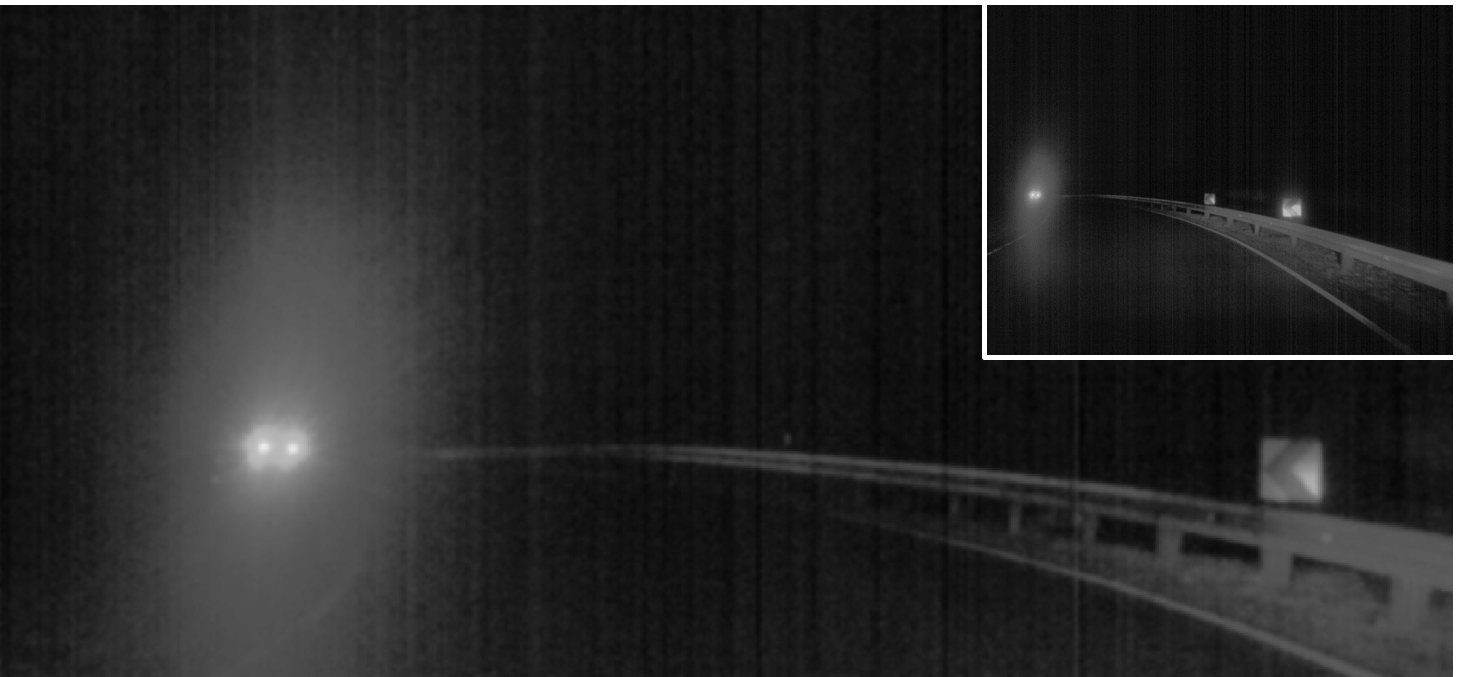}

\caption{\label{fig:opener-img}The three characteristic states how an oncoming
vehicle becomes visible. The images correspond to the scene 309 of
the PVDN dataset \parencite{Saralajew2021}. Each visualization shows
a cropped area of the entire image (shown in the upper right corner).
Left: First visible light artifacts at the guardrail (indirect light
instance). Middle: Vehicle becomes visible (direct and indirect light
instances). Right: Vehicle is visible (direct and indirect light instances).}
\end{figure*}
Another reason why computer vision systems cannot compete with humans
in general-purpose tasks is that computer vision systems are mostly
trained to solve one specific task. For this, the vision task is formulated
in a mathematical framework. For example, in object detection, the
most studied field in computer vision, the objects are marked by bounding
boxes, and the task is to predict and classify those bounding boxes
\parencite[e.\,g.,][]{Liu2019a}. Often, this sufficiently reflects
our human visual performance. But, in general, human visual perception
is more complex. Humans reason about the environment based on complex
learned causalities using \emph{all} the available information. If
we hear a siren, for instance, we expect the occurrence of an ambulance
and try to visually estimate the point of occurrence. Such causalities
to enrich our environmental model are present in everyday life: In
daylight, we use shadow movements and illumination changes to reason
about moving objects without having direct sight, and if we drive
a car through a village and see a ball rolling on the street, we expect
the occurrence of playing children. 

Another example for complex causalities to enrich the environmental
model happens while driving at night. At night, humans show impressive
abilities to foresee oncoming cars by analyzing illumination changes
in the environment like light reflections on guardrails (see left
image in \figref{opener-img}), a brightening of a turn ahead, or
unnatural glares in trees. Drivers use this provident information
to adapt their driving style proactively, for example, by turning
off the high beam in advance to avoid blinding of oncoming drivers
or by adapting their driving trajectory. In the scope of safe and
anticipatory driving, where time matters and the earlier information
is received the better it is, this human ability is obviously handy
and outperforms current computer vision systems used in vehicles.
\textcite{Oldenziel2020} analyzed this discrepancy between the human
detection capabilities and an in-production computer vision system
and quantified that humans are approximately $1.7$\,s faster. One
reason why state-of-the-art object detection systems are behind human
capabilities is that object detection systems rely on the assumption
that objects have clear, visible object boundaries. Even if this assumption
comes with a lot of advantages, like a well-defined description for
the enclosing bounding box of an object, it is not inherently applicable
to light artifacts---since usually light artifacts (illuminated areas)
have no clear object boundaries and the position of these light artifacts
does not directly correspond to the location of the light source.
Due to this assumption, the earliest point in time an oncoming vehicle
can be detected by state-of-the-art computer vision systems is after
almost full visibility (see the right image in \figref{opener-img}).

Nowadays, vehicles are increasingly equipped with driver assistance
systems, and manufacturers are working on self-driving cars. Therefore,
while driving, more and more tasks are controlled or supported by
systems such that the algorithms have more and more responsibility
to operate correctly in our complex environment. For safe and anticipatory
driving, time matters and $1.7$\,s are a non-negligible unexplored
potential to, for instance, plan driving trajectories, understand
the environment, or simply control the high beam to avoid glaring
of oncoming vehicles.

In this paper,\footnote{Note it is the continuation of earlier work \parencite{Oldenziel2020,Saralajew2021}
by our team with further improvements of the algorithm and the first
detailed presentation of the entire pipeline.} we study the task of detecting \emph{light artifacts} caused by the
headlights of oncoming vehicles so that we can reduce the aforementioned
time difference. To illustrate the usefulness (and to visualize the
detection results in the real world and in real time), a test car
is equipped with such a detection algorithm, and the information is
used to proactively control the glare-free beam system\footnote{A glare-free high beam system is an ADAS where the camera interacts
with a vehicle headlamp that consists of several adjustable pixels
like a video projector but with a coarse resolution. By processing
the information of detected vehicles of the driver assistance camera,
the goal of the headlamp is to illuminate the environment as much
as possible (adapted high beam) without blinding other drivers---pixels
that would illuminate other drivers are turned off.} \parencite[e.\,g.,][]{boke2015mercedes,Fleury2012,Kloppenburg2016}.
To this end, a full detection pipeline is implemented consisting of
\begin{enumerate}
\item the detection of light artifacts using the car's front camera,
\item the distance estimation to all detected light artifacts to provide
a three-dimensional localization, and 
\item the tracking of the light artifacts to perform a plausibility check
and to handle occlusions.
\end{enumerate}
Consequently, by detecting their light artifacts, it is possible to
detect that vehicles are oncoming before they are directly visible---this
means to \emph{providently detect oncoming vehicles}. In particular,
this investigation focuses on the first two points because state-of-the-art
approaches cannot be directly applied for the detection of light artifacts.
Consequently, on the computer vision level, our contributions are
\begin{itemize}
\item a fast and straightforward computer vision algorithm that is able
to detect light artifacts and
\item an investigation of methods to estimate the distance to light artifacts
to estimate their three-dimensional position.
\end{itemize}
On the system level, the contributions are
\begin{itemize}
\item the investigation of the toolchain to integrate such a provident vehicle
detection system in a vehicle to control an Advanced Driver Assistance
System (ADAS) and
\item the analysis of the time benefits that can be gained when using such
a detection system in vehicles.
\end{itemize}
The outline of the paper is as follows: First, in \secref{Terminology},
the terminology is presented that is used throughout the paper. Then,
in \secref{Related-Work}, the current state of the art in vehicle
detection at night is reviewed. This section also covers state-of-the-art
methods for distance estimation for camera-based systems. Based on
this, limitations of vehicle detection systems with respect to their
applicability for provident vehicle detection are highlighted in \secref{Inherent-limitation-of}.
In \secref{Method}, we present the developed provident vehicle detection
pipeline. To show the feasibility of such a system in a production
car environment, multiple experiments are performed in a real-world
environment. The experiments are described and evaluated in \secref{Experiments},
and \secref{Conclusion-and-Outlook} gives a conclusion and a future
outlook.

\section{Terminology\label{sec:Terminology}}

The following terminology is used:
\begin{description}
\item [{Light~artifact:}] Any form of artificial light in the image caused
by headlamps. This includes light reflections, glaring of areas above
a street, headlamp light cones.
\item [{Direct~light~instance:}] Light artifacts that are light sources
(direct view to the headlamp).
\item [{Indirect~light~instance:}] Light artifacts that are not direct
light instances like reflections on guardrails.
\item [{Provident~vehicle~detection:}] Detect the presence of oncoming
vehicles in an image (even if they are not in direct sight) by detecting
light artifacts caused by the vehicle's headlights.
\item [{Proposal~generation:}] An algorithm to extract relevant regions
(proposals) of an image in the form of bounding boxes. These regions
should contain light artifacts.
\item [{Proposal~classification:}] An algorithm to extract the proposals
that correspond to \emph{light artifacts} through classification.
\item [{Object~detector:}] The entire detector consisting of proposal
generation and proposal classification.
\end{description}

\section{Related work\label{sec:Related-Work}}

For autonomous driving and ADASs, the detection of vehicles is of
high priority to perform critical tasks like emergency braking maneuvers
or automatic high beam control. The commonly used sensor for that
is a driver assistance camera that captures images in the visual wavelength
range \parencite[e.\,g.,][]{Rezaei2017}. These images are then analyzed
to detect vehicles and to estimate the distance to detected vehicles
to determine the three-dimensional position. The following two paragraphs
present related work for these two topics (vehicle detection and distance
estimation) as they are key for our contributions. Additionally, a
final paragraph presents related work for provident vehicle detection.

\paragraph{Vehicle detection:}

The methods for vehicle detection by camera images depend on the visibility
conditions. For example, under good visibility conditions (e.\,g.,
daylight), vehicles are detected based on feature descriptors like
edge detectors and symmetry arguments using classifiers like support
vector machines on top \parencite[e.\,g.,][]{Sun2002,Sun2006,Teoh2011}
or are being detected by end-to-end trained deep NNs \parencite[e.\,g.,][]{Fan2016,Hassaballah2021,CarranzaGarcia2021}.
Under this condition, detectors often assume that vehicles can be
localized mainly by their contours, which is also supported by the
fact that the most commonly used annotation method for objects is
by bounding boxes that inherently require clearly visible object contours
to reliably annotate them (see the survey of \cite{Liu2019a}). 

If the visibility condition deteriorates, the detection performance
of the aforementioned methods decreases because of the reduced visibility
of object features such that specialized detection algorithms are
required. For instance, for adverse weather conditions like snow or
fog, \textcite{Hassaballah2021} proposed a promising image enhancement
strategy after which the aforementioned detectors can be applied,
and, for nighttime, researchers studied whether a style transformation
between nighttime and daylight images can be performed by a generative
adversarial network \parencite{Shao2021,Lin2021}. Even if the latter
idea is promising, it is not mature enough to compete with the performances
of specialized detection algorithms for nighttime (in terms of detection
rates and computational efficiency). 

At nighttime, due to low contrast, vehicles are usually detected by
locating their headlamps and rear lights singularities in the image
space caused by the luminous intensity of the light sources with rule-based
algorithms \parencite[e.\,g.,][]{Lopez.2008,P.F.Alcantarilla.2011,Eum.2013,Sevekar.2016,Pham2020}.
However, besides rule-based approaches, methods using NNs \parencite[e.\,g.,][]{Oldenziel2020,Mo2019,Bell2021}
or different imaging sensors like infrared cameras \parencite[e.\,g.,][]{Tehrani2014,Niknejad2011}
have been investigated as well. For vehicle detection at night, it
is difficult to judge which method is superior to another as the domain
has not agreed on a benchmark dataset like the KITTI benchmark for
daylight \parencite{Geiger2012}, which is also criticized by other
authors \parencite{Sun2006,Juric.2014}---authors reported results
with around 90\,\% accuracy and small error rates for both rule-based
and NN-based classifiers on their \emph{private} datasets (e.\,g.,
compare the evaluations of \cite{Mo2019} and \cite{Satzoda2019}).
Nevertheless, it has to be expected that rule-based methods are superior
if computational complexity constraints apply (e.\,g., see the number
of parameters and the number of GFLOPs in Table~2 of \cite{Saralajew2021})
and if somebody wants to use the detections for high-stakes decisions
\parencite{Rudin2019}. Therefore, as the scope of this work is to
apply the detection algorithm in a test vehicle for a driver assistance
system, we focus on a rule-based proposal generation algorithm with
a shallow NN on top to classify the proposals (whereas the NN is not
specific to our approach and can be replaced by any other classification
method).

\paragraph{Distance estimation:}

As we consider the application of the detection pipeline in a test
car to realize a prototypical customer functionality, requirements
by law have to be considered. Assuming the usage of the detected vehicle
information to control the vehicle's high beams at night, the distance
at which a visible vehicle must be detected is regulated by the respective
\textcite{UNECERegulation2016} to avoid blinding of other drivers:
``The sensor system shall be able to detect on a straight level road:
(a) An oncoming power driven vehicle at a distance extending to at
least 400\,m; (b) A preceding power driven vehicle or a vehicle-trailers
combination at a distance extending to at least 100\,m'' (§\,6.1.9.3.1.2).

In order to ensure such large detection distances, researchers who
developed the detection pipeline for automotive use cases rely primarily
on the distance estimation by a ground plane assumption \parencite[e.\,g.,][]{P.F.Alcantarilla.2011,Juric.2014,Eum.2013,Chen2008,Kuo2010,Schamm2010}:\footnote{Note that other sensors like radar and LiDAR (even if progressing
constantly) cannot provide a reliable distance estimation up to such
large distances (e.\,g., see the range specifications by \cite{Kukkala2018}).
In addition, a single sensor solution is also preferred in the context
of low-cost solutions.} Knowing the extrinsic and intrinsic parameters of the camera mounted
in the vehicle and assuming or estimating the ground ``plane'' in
front of the vehicle, the distance to vehicles can be estimated. Using
this technique, \textcite{P.F.Alcantarilla.2011} reported detection
distances of up to 700\,m for oncoming and 200\,m for preceding
vehicles. Other authors purposefully used the known calibration of
the camera and different assumptions like a known distance between
headlamp pairs or vanishing point estimation to estimate the distance
to vehicles \parencite[e.\,g.,][]{Chen2012,Chen2009}.

In addition to the methods mentioned before, researchers investigated
single image depth estimation approaches by the use of deep NNs \parencite[e.\,g.,][]{eigen2014,Laina2016},
where the idea is that deep NNs can learn to contextualize a scene
and the arrangement of objects to the depth information. Often these
methods only return relative depth information that is only accurate
in the close range so that it might not be applicable for our purpose.
Another approach is depth estimation by structure from motion \parencite[e.\,g.,][]{furukawa2004,saponaro2014,gallardo2017},
where the structure (depth) is estimated by analyzing the movement
of objects. In principle, this is similar to the distance estimation
by stereo vision systems \parencite{hamzah2016} because, in both
approaches, correspondences between images have to be found and analyzed.
However, the latter is not applicable for our use case since we focus
on monocular camera systems. Finally, in general, several of the mentioned
concepts are used in depth estimation from video \parencite[e.\,g.,][]{Zhou2017,ranftl2016,gordon2019},
where the goal is to provide an accurate distance estimate by analyzing
consecutive images of a video. 

In summary, all these methods have pros and cons and require certain
assumptions to be valid. Therefore, we discuss their applicability
for our use case in the following sections and, additionally, investigate
a use case specific approach that uses predictive street data in order
to locate light artifacts.

\paragraph{Provident vehicle detection:}

The provident detection of objects was already studied by other authors:
In daylight, \textcite{FelixMaximilianNaser.18.01.2019} providently
detected objects by analyzing shadow movements; At nighttime, \textcite{Oldenziel2020}
and \textcite{Saralajew2021} studied the task to providently detect
oncoming vehicles by detecting light artifacts produced by their headlights.

\textcite{Oldenziel2020} analyzed the discrepancy between the human
abilities and an in-production computer vision system in detecting
oncoming vehicles. Notably, based on the results of a test group study,
the authors specified the deficit in detecting oncoming vehicles providently
by 1.7\,s on average in favor of humans. Since this is a significant
amount of time, the authors studied whether it is possible to detect
oncoming vehicles based on light artifacts by training a Faster-RCNN
architecture \parencite{fasterrcnn} on a small private dataset annotated
by bounding boxes. The presented results showed that the NN learned
the task to some extent. However, the analysis of the detection results
raised concerns whether an annotation method with bounding boxes (even
if most commonly used) is a good annotation scheme for light artifacts
due to a high annotation uncertainty because of unclear object boundaries---light
artifacts are fuzzy and of weak intensity such that clear boundaries
are missing. These results are partly orthogonal to \textcite{Bell2021},
where the authors annotated vehicles in nighttime images of traffic
surveillance cameras with \emph{keypoints} because the blurry edges
of the vehicles due to motion blur and the saturated pixels due to
the bright light cones of vehicles headlamps did not allow a reliable
annotation of the vehicles by bounding boxes.

The annotation by keypoints causes the difficulty that the majority
of state-of-the-art object detectors cannot be applied because they
require bounding boxes \parencite[e.\,g.,][]{Liu2019a}. For this
reason, \textcite{Bell2021} used foveal classifiers (a set of classifiers
where each classifier is trained to provide the classification result
for a particular image region). The disadvantage of this approach
is that the localization performance depends on the number of classifiers
and their distribution over the image. For their use case (traffic
surveillance), however, this approach is feasible and provides good
results. But for the application of provident vehicle detection, the
limited localization performance and the dynamics of oncoming vehicle
scenarios are expected to negatively affect the applicability of this
detection approach. To overcome this and to apply state-of-the-art
object detectors, the authors proposed a simple transformation by
sampling bounding boxes of random size around each keypoint in order
to derive bounding boxes. Based on this, they trained a YOLOv3 \parencite{Redmon2018}
and a Faster R-CNN network, but these networks scored worse than their
foveal classifier in the presented evaluation. Whether this result
is partly due to the simplicity of their bounding box generation,
perhaps causing unexpected biases, is unclear.

\textcite{Saralajew2021} extended the work of \textcite{Oldenziel2020}
and published the PVDN dataset, the first containing approximately
60\,K driver assistance camera images (grayscale) annotated by keypoints
for the task to providently detect oncoming vehicles at nighttime.
Together with the dataset published by \textcite{Bell2021}, these
two datasets are the only large-scale datasets publicly available
for the detection of vehicles at nighttime and annotated by keypoints---other
\emph{available} datasets for this task use bounding boxes \parencite[e.\,g.,][]{Rezaei2017,Duan2018}
or masks \parencite{Rapson2018}. By using keypoints for the annotation
of light artifacts, \textcite{Saralajew2021} presented an approach
to reliably annotate light reflections, which are so fuzzy and weak
in intensity that they cannot be objectively annotated by bounding
boxes (as concluded by \cite{Oldenziel2020} and showed in a test
by \cite{Saralajew2021}). Similar to \textcite{Bell2021}, using
the keypoints as initial seeds, the authors further explored methods
to extend the keypoint annotations to bounding boxes with low annotation
uncertainty so that state-of-the-art object detection methods can
be applied. To this end, they trained several machine learning algorithms
for the task of detecting light artifacts. The two types of architectures
used for this experiment are YOLOv5 networks\footnote{\url{https://github.com/ultralytics/yolov5}}
and a two-phase algorithm consisting of a rule-based blob detector
followed by a shallow NN\@. Both methods show promising results and
provide a strong baseline for further experiments. As mentioned earlier,
we build on the latter and fine-tune the architecture such that a
new benchmark is achieved.

\section{Inherent limitation of current systems\label{sec:Inherent-limitation-of}}

\begin{figure*}
\begin{centering}
\includegraphics[scale=0.95]{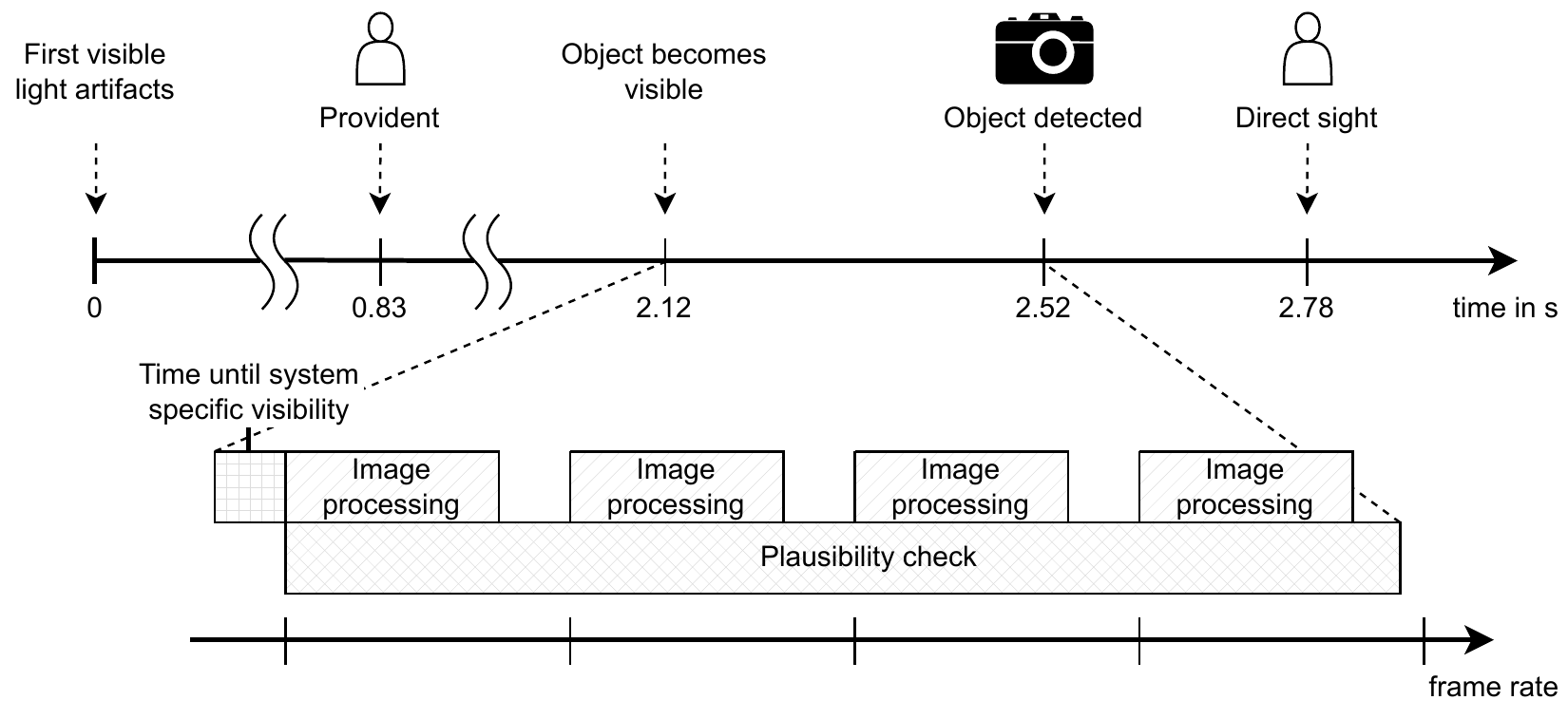}
\par\end{centering}
\begin{centering}
\caption{Visualization of the timings of how an in-production system and humans
(provident and after direct sight) perceive oncoming vehicles at night.
The times are estimates from a test group study performed by \textcite{Oldenziel2020}
and are (of course) dependent on the scenario. However, they illustrate
the inherent discrepancy. Additionally, the system-related latency
between \textquotedblleft object becomes visible\textquotedblright{}
and \textquotedblleft object detected\textquotedblright{} is qualitatively
split into the single steps.\label{fig:motivation}}
\par\end{centering}
\end{figure*}
Simply said, the motivation of this work is to provide the information
about oncoming vehicles at night earlier than current systems do---in
the best case before they are directly visible---to ensure safe and
anticipatory driving. Currently, there is a technical limitation in
current systems regarding how early they can perceive a vehicle (see
\figref{motivation}), caused by the commonly used object detection
paradigms and the system-related latencies. Within this section, we
explain why these limitations exist and are inherent. Knowing these
limitations is essential to understand what can be achieved with the
presented approach (how fast can vehicles be detected).

\subsection{Object detection paradigms\label{subsec:Object-detection-paradigms}}

First, it must be noted that current camera-based perception models
used to detect vehicles at night are object detectors. As the most
reliable information source, headlamps of other vehicles are used
to detect the position of vehicles. Consequently, headlamps are used
as ``objects'' from which succeeding systems can infer the location
of vehicles present in the image \parencite[e.\,g.,][]{P.F.Alcantarilla.2011}.
While being a robust reference, the restriction on headlamps limits
the performance of such systems, since the earliest point in time
they can perceive a vehicle is when they have direct sight to the
vehicle (see ``object becomes visible'' in \figref{motivation}
and the middle image in \figref{opener-img}). As already mentioned
in \secref{Introduction}, this differs from how humans estimate whether
and where a vehicle is oncoming because humans can react to light
artifacts like the light reflections on the guardrail in \figref{opener-img}.
Thus, the question is why light artifacts are not naturally being
detected or tried to being detected by current vehicle detection systems
considering the apparent discrepancy between humans and systems regarding
this task (see the time gap between the human provident and camera-based
object detection in \figref{motivation}). We can only speculate why
this is the case but expect that one reason is the object detection
paradigm: The algorithms detect objects that match an object definition.
For example, if the object detector is a bounding box regressor, it
must be possible to specify the object boundaries to define bounding
boxes. In vehicle detection at night, it is possible to apply this
strategy for headlamps (direct light instances) as they cause intensity
singularities in the image (extraordinary high-intensity peaks) with
sharp gradients. However, for light reflections (indirect light instances),
this strategy is not appropriate since they often illuminate almost
homogeneously larger areas with small gradients, and their intensity
varies heavily depending on their strength and other global light
sources. Therefore, indirect light instances cannot be treated as
objects without further thoughts due to their unclear object boundaries,
causing annotation difficulties. Building on the work of \textcite{Saralajew2021},
we tackle this challenge by using keypoint annotations to derive a
suitable object detector for detecting all sorts of light artifacts.

\subsection{System-related latencies\label{subsec:System-related-latencies}}

As already mentioned in \secref{Related-Work}, \textcite{Oldenziel2020}
presented the results of a test group study that investigated the
detection latency of an in-production computer vision system and humans---the
results are summarized in \figref{motivation}. In particular, they
observed that even if, in theory, detection systems are able to detect
vehicles directly after direct sight, on average, they have a system-related
latency. This latency is caused by a small delay until the object
paradigm is fulfilled, by the image processing steps of the detection
pipeline (as described in \secref{Introduction}), and by the time
needed to perform the plausibility check. In summary, the following
steps cause the latency in a vehicle detection system---which is
also qualitatively visualized in \figref{motivation}:
\begin{enumerate}
\item After the vehicle starts to become visible, there is a system-specific
time until the vehicle has a visibility status that fulfills the object
definition. After that, the object can be potentially detected by
the computer vision system (compare the middle and right image in
\figref{opener-img}).
\item If an image is captured with an object that fulfills the object definition,
a latency in the detection is caused by the image processing time
(object detection, distance estimation, and tracking). This latency
is lower than the frame rate of the camera.
\item Finally, the plausibility check causes a latency of several frames
due to semantic and \emph{temporal} object validations. This step
is often performed in combination with the object tracking in order
to safely predict there is an oncoming vehicle.
\end{enumerate}
This latency is frequently discussed in combination with glare-free
high beam systems because of the possible caused glare \parencite[e.\,g.,][Chapter 3.3.1]{Helmer2021}:
\textcite[Chapter 5.4.3]{Hummel2009} reported that the plausibility
check causes a latency between two and four frames, where a frame
has a processing time of $\unit[45]{ms}$, and measured a delay from
capturing the image until the high beam adapted of $\unit[\left(196\pm21\right)\!]{ms}$.\footnote{Unless otherwise specified, expressions such as $196\pm21$ represent
a mean of 196 with a standard deviation of 21.} \textcite[Chapter 3.2.4]{Totzauer2013} reported a similar result
of $\unit[\left(288\pm17.4\right)\!]{ms}$ total system delay. Moreover,
\textcite{Lopez.2008} reported a validation time of two or three
frames for oncoming vehicles. Unfortunately, they do not specify the
camera's frame rate, so these results cannot be converted into seconds.
However, they state that the overall processing delay is less than
$\unit[200]{ms}$. The approach of \textcite{P.F.Alcantarilla.2011}
validates detected vehicles for five frames at a frame rate of $30$,
leading to a minimum delay of around $\unit[167]{ms}$. Summarily,
current camera-based systems for detecting vehicles at night are expected
to have an inherent detection delay caused by the mandatory vehicle
validation procedure and the assumption that vehicles are characterized
solely by headlights (as discussed in the previous \subsecref{Object-detection-paradigms}).
Consequently, if such a vehicle detection system is used to deploy
a glare-free high beam system, it has to be expected that oncoming
drivers are exposed to high beam light patterns when they appear in
direct sight. Whether these glare moments are critical has not yet
been conclusively clarified \parencite{Helmer2021}.

Not only object detection systems have an internal latency, but also
humans: reaction time. In \figref{motivation}, the human reaction
times during the test group study are illustrated. As \textcite{Oldenziel2020}
showed, the camera-based vehicle detection is approximately 200\,ms
faster in a fair setting (allowed detection after direct sight) than
its human counterpart. However, human detection almost reaches the
minimal possible detection time, acting only approximately 800\,ms
after the first indication of oncoming vehicles (compare with the
reaction times for braking maneuvers \cite{Green2000}). 

In \secref{Method}, we present a detection system that is able to
detect light artifacts. Even if this system reacts to light artifacts,
it still has the inherent system-related latency. Therefore, depending
on the scenario, it does not necessarily detect oncoming vehicles
before direct sight but shifts as much as possible of the inherent
latency before the moment of direct sight and, thus, detects oncoming
vehicles faster than current systems do.

\section{Method\label{sec:Method}}

In this section, the methodology for the detection of light artifacts
is presented. First, an object detector is proposed that can detect
light artifacts. After that, different techniques to estimate the
distance are described. Such a distance estimation is needed to locate
the detected objects in the three-dimensional space. Finally, a tracking
algorithm is outlined to stabilize the detections and to perform a
plausibility check.

\subsection{Object detector\label{subsec:Object-Detector}}

\begin{figure*}
\centering{}\includegraphics{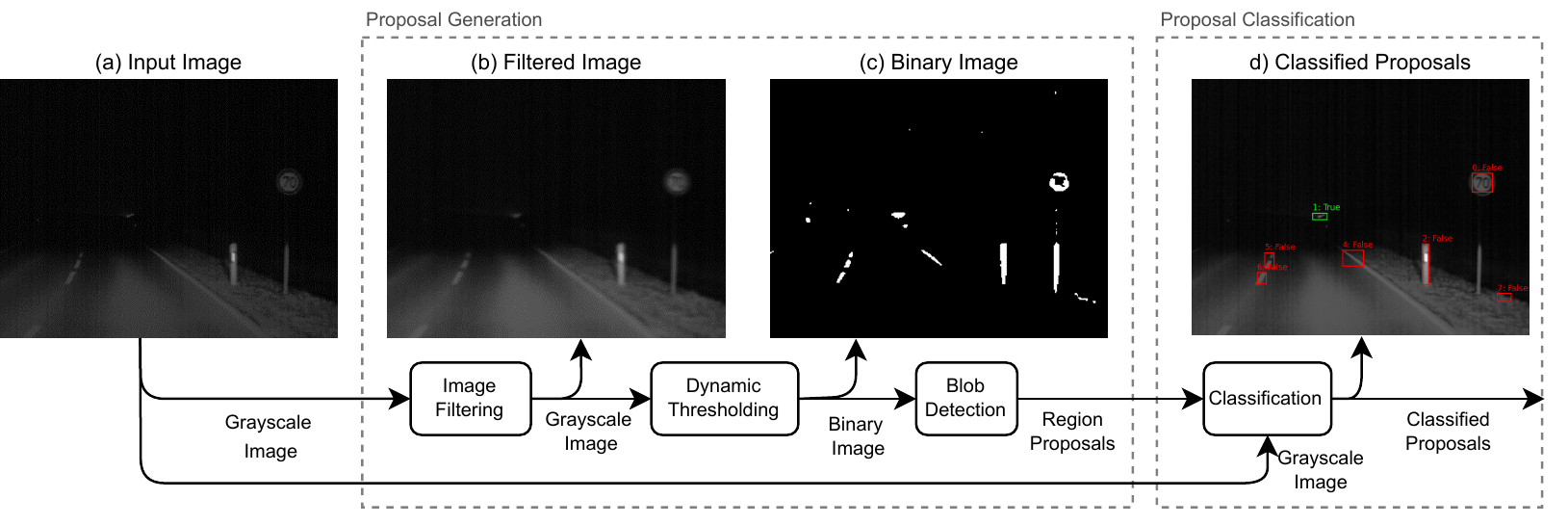}\caption{Overall object detector pipeline. First, region proposals for light
artifacts are generated using a custom approach described by \textcite{Saralajew2021}
(\textquotedblleft Proposal Generation\textquotedblright ). The region
proposals are then passed to a classifier (\textquotedblleft Proposal
Classification\textquotedblright ) for the binary classification to
\textquotedblleft light artifact\textquotedblright{} and \textquotedblleft non-light
artifact.\textquotedblright{} For both steps, modules, as well as intermediate
results (images), are shown. Image (a) shows the raw input image,
(b) the filtered image, (c) the binary image inferred through the
adaptive thresholding procedure (white: 1; black: 0), (d) the classified
proposals (green: light artifact, red: non-light artifact). The image
used for this figure corresponds to a crop of image number 98\,768
of sequence 286 in the PVDN dataset and shows the detection of an
indirect light instance.\label{fig:Object-detector-pipeline:}}
\end{figure*}
The first element in the vehicle detection pipeline is the object
detector. The task here is to detect both direct and indirect light
instances within the camera image. The feasibility of the light artifact
detection was shown by \textcite{Oldenziel2020} through multiple
practical examples. The general setup for such a detector can be divided
into the following sub-tasks:
\begin{enumerate}
\item Generate region proposals based on local features;
\item Classify the proposals to reduce the amount of false-positive detections.
\end{enumerate}
This pipeline is used in many state-of-the-art systems as well as
in machine learning object detectors \parencite[e.\,g.,][]{fasterrcnn}.
The usual approach for such a system would be to use an NN-based system
in an end-to-end manner. As already said, we rejected this approach
because of its inevitable obscure nature and computational load. Instead,
the method proposed by \textcite{Saralajew2021} with a tailored region
proposal generation algorithm and an NN classifier is used. The resulting
overall pipeline is depicted in \figref{Object-detector-pipeline:}.
In the following paragraphs, each module of the object detector is
described in more detail.

\paragraph{Pipeline:}

First, a dynamic thresholding procedure is performed to retrieve intensity
regions of interest from the image. Bounding box proposals are then
inferred through a blob detection in the generated binary image (1:
above threshold; 0: below threshold). The classification is performed
using a small Convolutional NN (CNN). The results are bounding box
representations of light artifacts within the image. 

Image preparation---The raw image is filtered to reduce the amount
of noise present. First, the image size is decreased to half (640$\times$480
pixels) by a bilinear interpolation in order to suppress small noisy
image regions (it also reduces the computational complexity of the
later steps). Second, noise is further removed by applying a Gaussian
blur over the entire image. This smooths out edges and removes high-frequency
noise like salt-and-pepper noise. The effects of the filtering are
depicted in \figref{Object-detector-pipeline:}b.

Dynamic thresholding---Due to the low intensities for light reflections
and glares, a global thresholding strategy is not suitable to retrieve
interesting regions from the raw image. In contrast, all considered
artifacts share the common feature of a higher intensity relative
to their surroundings \parencite{Saralajew2021}. This can be used
to perform dynamic thresholding on the image. Therefore, a pixel-wise
threshold is calculated to retrieve interesting regions. 

The criterion for the dynamic threshold $T(x,y)$ at pixel $(x,y)$
is defined as the following: 
\begin{equation}
T(x,y)=\mu(x,y)\cdot\left(1+\kappa\cdot\left(1-\frac{\Delta(x,y)}{1-\Delta(x,y)}\right)\right),\label{eq:dynthresh}
\end{equation}
with $\mu(x,y)$ being the local mean intensity---calculated over
a fixed-sized window $w$ around the pixel $(x,y)$---and $\Delta(x,y)=I(x,y)-\mu(x,y)$
being the deviation of the pixel intensity $I(x,y)$ from the local
mean. The sensitivity of this threshold can be adjusted using the
factor $\kappa\in\mathbb{R}$. \Eqref{dynthresh} is adapted from
\textcite{singh2011thresholding}, who originally developed this technique
to binarize documents. Comparisons with other threshold techniques
showed that this method yields high-quality results. Also, by using
the integral image to compute the local means, this method can be
efficiently implemented \parencite{singh2011thresholding}. The threshold
$T(x,y)$ is calculated for every pixel in the filtered image and
used to infer a binary image $B(x,y)$ with
\begin{equation}
B(x,y)=\begin{cases}
1 & \text{if }I(x,y)>T(x,y),\\
0 & \text{otherwise.}
\end{cases}
\end{equation}
An example of such a binary image is shown in \figref{Object-detector-pipeline:}c. 

Blob detector---The binary image contains multiple, unconnected regions
generated by the thresholding step. For ease of use and further handling,
these regions are compressed into bounding boxes. This is achieved
by applying a standard blob detection routine to find connected areas
and allowing gaps of the size $d$---measured with respect to the
$L_{\infty}$ distance.\footnote{The $L_{\infty}$ distance, also denoted as the Chebyshev distance,
between two vectors $\mathbf{x}$ and $\mathbf{y}$ is the maximum
absolute deviation in any dimension: $L_{\infty}\left(\mathbf{x},\mathbf{y}\right)=\max_{i}\left|x_{i}-y_{i}\right|$.} After the bounding boxes have been computed, they are filtered by
removing bounding boxes where the mean absolute deviation of the included
intensity values is smaller than a threshold~$s$.

Classification---The bounding boxes still contain many false positives
because, simply put, all bright areas of the image are detected. This
allows for a high recall of interesting regions but also yields a
low precision and therefore reduces the quality of following modules
(e.\,g., the glare-free high beam functionality). Therefore, a shallow
NN is added to classify each of the proposals (this strategy is similar
to a Faster-RCNN architecture). For that, to provide context information
for each bounding box, an \emph{enlarged} region around each proposal
bounding box is passed through a CNN\@. The network classifies the
proposal to be either true positive or false positive (see also \subsecref{Object-detector}
for a formal definition of true and false positive), leading to a
binary classification problem. Here, a proposal is considered true
positive if it coincides with a light artifact (direct or indirect)
of oncoming vehicles. Therefore, false-positive proposals are all
remaining regions. The reason for posing this as a binary classification
problem is explained in more detail in \subsecref{Experimental-platform}.
As the network architecture is equivalent to \textcite{Saralajew2021},
we will not discuss the classifier architecture in detail. For more
information, see \textcite{Saralajew2021} or the publicly available
implementation in the corresponding GitHub repository.

\paragraph{Efficiency:}

The approach used to detect light artifacts was chosen to allow for
a suitable implementation in a production car, where only limited
computational resources are available. While many detection and recognition
systems designed for the automotive context rely heavily on parallel
computing (e.\,g., through GPUs, TPUs), such hardware is not yet
implemented in most production cars, limiting the practical usage
of these systems. Even if more and more computational power is available
in the upcoming years, resources will always be limited as the number
of functions increases as well. Therefore, two of the major requirements
for the detection system are to be computationally efficient and to
not rely too heavily on additional hardware. The simple operations
used to build the proposal generation are a result of these requirements.
The classification is still performed on a GPU but is still efficient
enough to be implemented on a production car's hardware with only
minor adjustments. As shown in literature, end-to-end learned systems
outperform conventional methods like the proposed one. This is also
partly true for our case when computational resources are unlimited.
However, this fact changes with the constraint of limited computational
resources, as shown by the evaluation in \subsecref{Object-detector}.

\subsection{Distance estimator\label{subsec:Distance-Estimator}}

In real-world driving scenarios, it is often not sufficient to just
provide the spatial information of the detected object in the two-dimensional
image space. Only knowing \emph{where} the object of interest is located
in the environment enables the vehicle to react appropriately---for
example, for performing emergency brakes or adjusting the adaptive
high beams. Therefore, it is necessary to compute an estimate for
the three-dimensional position of detected light artifacts.

\begin{table*}
\centering{}\caption{Summary of methods for monocular visual distance estimation (three-dimensional
object localization). All methods are analyzed and assessed regarding
their applicability to low-texture and dark images, computational
complexity, applicability to arbitrarily deforming objects (such as
light artifacts), and large distances according to the requirements
discussed in \secref{Related-Work}.\label{tab:summary-depth-estimation}}
\begin{tabular*}{1\textwidth}{@{\extracolsep{\fill}}>{\centering}m{0.3\textwidth}>{\centering}m{0.13\textwidth}>{\centering}m{0.13\textwidth}>{\centering}m{0.13\textwidth}>{\centering}m{0.1\textwidth}}
\toprule 
Method & Low-textured,\\
dark image & Arbitrarily de-\\
forming objects & Computational\\
complexity & Large\\
distances\tabularnewline
\midrule
\midrule 
Monocular, single image depth estimation \parencite[e.\,g.,][]{wofk2019,Laina2016,eigen2014} & no & yes & high & no\tabularnewline
\addlinespace
Depth estimation from video \parencite[e.\,g.,][]{Zhou2017,ranftl2016,gordon2019} & no & no & high & no\tabularnewline
\addlinespace
Structure from motion \parencite[e.\,g.,][]{furukawa2004,saponaro2014,gallardo2017} & partly & partly & high & no\tabularnewline
\addlinespace
Ground plane approaches \parencite[e.\,g.,][]{P.F.Alcantarilla.2011,Eum.2013,Juric.2014} & yes & yes & low & yes\tabularnewline
\bottomrule
\end{tabular*}
\end{table*}
As discussed in the related work in \secref{Related-Work}, in the
literature, there are several methods described to perform the distance
estimation\footnote{Also referred to as object localization or depth estimation.}
to locate objects. However, the special use case of nighttime images
captured by a monocular grayscale camera adds clear restrictions.
The general problem is that the images are relatively dark and low-textured
(e.\,g., see \figref{opener-img}), which complicates the application
of state-of-the-art depth estimation methods. Also, light reflections
can be considered non-rigid, arbitrarily deforming objects over time.
In addition, the overall goal of running the method in real-time places
a constraint on computational complexity, and the goal of using it
for a real-world use case requires a certain range.

\tabref{summary-depth-estimation} presents a summary of possible
applicable distance estimation methods. Due to the aforementioned
constraints, the applicability of depth estimation from video and
structure from motion is not possible. Additionally, monocular, single
image depth estimation approaches have a too high computational complexity
and cannot provide the required range, so they cannot be used for
the studied approach as well.\footnote{Even if these methods are not a good choice for the studied use case
with respect to the computational complexity, some of them were tested
in a proof-of-concept investigation on grayscale daylight and nighttime
images. The results were not satisfying, which underlined the exclusion
from further studies.} Hence, the only applicable state-of-the-art method is the object
localization through ground plane approaches. Note that this method
assumes that \emph{an object is located on the ground plane}, which
is not necessarily true for light artifacts (e.\,g., a light artifact
on a guardrail). In previous work about vehicle detection where the
vehicle's headlamps are detected \parencite[e.\,g.,][]{P.F.Alcantarilla.2011},
the fulfillment of this assumption is achieved by assuming a fixed,
known height of the headlamps and by shifting the ground plane by
this height.

\begin{figure*}
\centering{}\hspace*{\fill}\textbf{}\subfloat[Projection of the PSD road-graph into the image.]{\begin{centering}
\includegraphics[height=5cm]{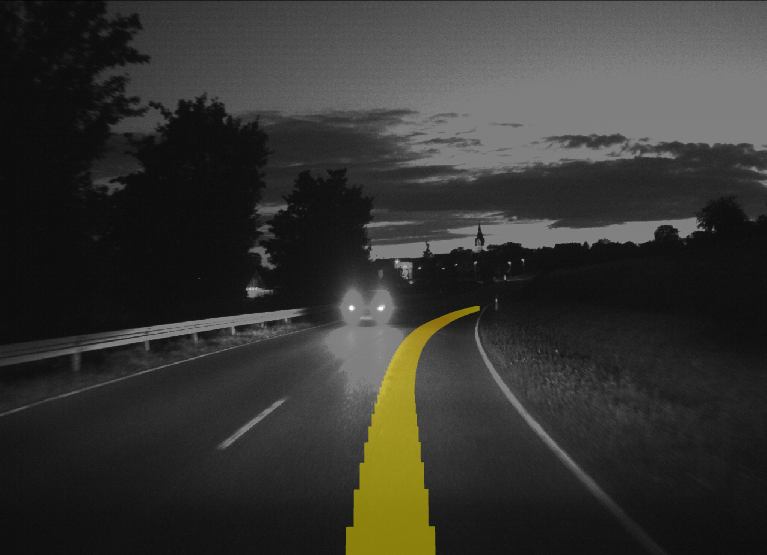}
\par\end{centering}
\textbf{}}\hspace*{\fill}\subfloat[Visualization of the PSD road-graph in the three-dimensional vehicle
coordinate frame.]{\begin{centering}
\includegraphics[width=6.909cm,height=5cm]{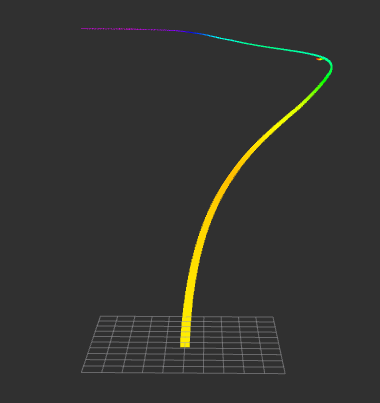}
\par\end{centering}
}\hspace*{\fill}\caption{Visualization of the road geometry obtained from the PSD.\label{fig:psd-visualization}}
\end{figure*}
\begin{figure}
\begin{centering}
\includegraphics[scale=0.7]{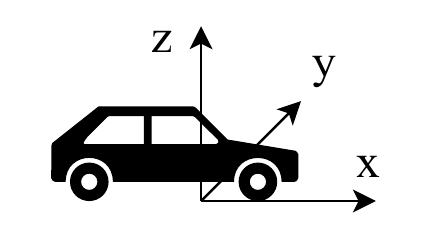}
\par\end{centering}
\caption{Vehicle coordinate system.\label{fig:Vehicle-coordinate-system}}

\end{figure}

To overcome the limitations of the methods listed in \tabref{summary-depth-estimation},
we also evaluated a rather unconventional method for estimating the
distance of light artifacts by fusing the position of the object in
the image with Predictive Street Data (PSD). The PSD protocol contains
information about the road geometry ahead of the vehicle (see \figref{psd-visualization})
based on map data and GPS and is used, for instance, for advanced
navigation functionalities or adaptive cruise control. With this data,
the road lying ahead can be projected into the vehicle coordinate
system (see \figref{Vehicle-coordinate-system}), giving a three-dimensional
representation of the road geometry. For our implementation, the road
ahead described by the PSD is defined as a set of $n$ points $\mathcal{P}=\left\{ P_{0},P_{1},P_{2},\dots,P_{n}\right\} $,
where $P_{i}$ is a point in the vehicle coordinate system lying on
the road ahead described by the PSD\@. A point was sampled for every
meter. At the same time, knowing the intrinsic and extrinsic camera
calibration,\footnote{For in-production vehicles with a series driver assistance camera,
these parameters are known and are dynamically corrected to account
for vehicle dynamics and so on.} a point $\left(x,y\right)$ in the image can be associated with a
camera ray $\vec{\mathbf{x}}\left(t\right)$ in the vehicle coordinate
system \parencite[e.\,g., see][]{hartley2003}. Assuming that a detected
light artifact always lies on or at least close to the road, the ray
$\vec{\mathbf{x}}\left(t\right)$ and the road ahead described by
$\mathcal{P}$ are used to search for the closest point $P_{i}\in\mathcal{P}$
with respect to $\vec{\mathbf{x}}\left(t\right)$, which is an intersection
between the road ahead and the ray in the best case. This point is
then considered as the object position in the vehicle coordinate system.
The distance between a point $P_{i}$ and the ray $\vec{\mathbf{x}}\left(t\right)$
can be calculated by 
\begin{equation}
D\left(\vec{\mathbf{x}}\left(t\right),P_{i}\right)=\frac{\left\Vert \left(\vec{\mathbf{p}}_{i}-\vec{\mathbf{a}}\right)\times\vec{\mathbf{n}}_{Q}\right\Vert }{\left\Vert \vec{\mathbf{n}}\right\Vert },
\end{equation}
where $\vec{\mathbf{n}}$ and $\vec{\mathbf{a}}$ are the parameter
vectors of the ray $\vec{\mathbf{x}}\left(t\right)=\vec{\mathbf{a}}+t\vec{\mathbf{n}}$,
and where $\vec{\mathbf{p}}_{i}$ is the canonical vector representation
of the point $P_{i}$. With this, the closest point of the road ahead
is given by 
\begin{equation}
P^{*}=\arg\min\left\{ D\left(\vec{\mathbf{x}}\left(t\right),P_{i}\right):P_{i}\in\mathcal{P}\right\} ,
\end{equation}
which determines the position of the object in the vehicle coordinate
system---detected in the image at position $\left(x,y\right)$.

Based on the PSD data and the available ground plane approaches, the
following four methods are analyzed in the experiments:
\begin{description}
\item [{PSD-3D}] uses the PSD road geometry (three-dimen\-sional) and
searches for the closest point along a camera ray of the detected
object.
\item [{PSD-3D\nolinebreak[4]\hspace{-.05em}\raisebox{.4ex}{\tiny\bf +}}] follows
the PSD-3D principle but corrects the vehicle orientation (yaw angle
and lateral offset) by road markers detected with the camera.\footnote{The vehicle orientation is calculated using the onboard, in-production
vehicle orientation algorithm.}
\item [{PSD-2D}] follows the PSD-3D principle but simplifies the problem
to a two-dimensional coordinate system by ignoring the elevation information.
\item [{GP}] simply assuming the road ahead as a plane (the so-called ground
plane) and computing the intersection of the object's associated camera
ray $\vec{\mathbf{x}}\left(t\right)$ with the plane \parencite[e.\,g.,][]{Juric.2014}.
In symbols, the ground plane is assumed to be the plane defined by
$z=0$ (see the vehicle coordinate system in \figref{Vehicle-coordinate-system})
so that the intersection with the camera ray $\vec{\mathbf{x}}\left(t\right)$
results from the $z$-components of the vectors $\vec{\mathbf{a}}$
and $\vec{\mathbf{n}}$ at the position $\vec{\mathbf{x}}\left(\frac{-a_{z}}{n_{z}}\right)$,
which determines the three-dimensional position of the object.\footnote{Note that researchers also investigate how to improve the ground plane
assumption to improve the distance estimation \parencite[e.\,g.,][]{P.F.Alcantarilla.2011}.
Therefore, this simple method can be further improved.}
\end{description}

\subsection{Object tracking\label{subsec:Object-tracking}}

The object detection and distance estimation are frame-based computations
and, therefore, can be unstable with respect to the temporal context
(e.\,g., if a vehicle gets occluded). To improve the detection stability,
object tracking algorithms are used
\begin{itemize}
\item to match the objects between different frames,
\item to predict the position of occluded objects, and
\item to increase the precision of the vehicle detection.
\end{itemize}
In the literature, the dominating algorithms for this are $\alpha$-$\beta$
filters \parencite[e.\,g.,][]{Pham2020,Lopez.2008} and Kalman filters
\parencite[e.\,g.,][]{P.F.Alcantarilla.2011,Teoh2011}, whereas the
former is a derivative of the latter. Due to the computational efficiency
of the $\alpha$-$\beta$ filter and because good estimates for the
noise covariance matrices to instantiate a Kalman filter are not known
(especially for light artifacts), the proposed tracker is mainly composed
of $\alpha$-$\beta$ filters:
\begin{itemize}
\item $\alpha$-$\beta$ filter in the two-dimensional image space to predict
and estimate the position of bounding boxes;
\item $\alpha$-$\beta$ filter to predict and estimate the distances to
the objects;
\item moving mean filter to estimate the confidence.
\end{itemize}
Between different frames, the object matching is performed by computing
the intersection-over-union between the tracked objects and the detected
objects and assigning the detected objects to the tracked objects
with the highest intersection-over-union. To handle noise in the detections
with respect to the bounding box size, the bounding box size of the
detected objects is slightly increased before the intersection over
union is computed.

If an object is occluded (not detected in the last frame), the prediction
of the $\alpha$-$\beta$ filter is used to forecast the position
of the object for a maximal number of three frames before it is removed
from the list of tracked objects. Additionally, to increase the precision
of the vehicle detection system, an object is only output when it
is already detected for a minimal number of five frames and if the
estimated confidence is greater than a threshold of 0.5---thus, the
tracker also operates as a \emph{plausibility checker}, which is a
common strategy as discussed in \subsecref{System-related-latencies}.
Finally, to lower the number of tracked objects, the tracker only
considers objects with a confidence value greater than 0.1.

\section{Experiments\label{sec:Experiments}}

The experiments described in this section aim
\begin{itemize}
\item to optimize the baseline bounding box annotation quality and, therefore,
the detector performance presented by \textcite{Saralajew2021},
\item to evaluate the distance estimation methods,
\item to quantify the time benefit of the proposed system in terms of a
provident detection of oncoming vehicles with respect to both human
performance and an in-production computer vision system for vehicle
detection at night, and 
\item to demonstrate the utility of the provident vehicle detection information
by integrating the proposed detection system into a test car and realizing
a glare-free high beam functionality. 
\end{itemize}
In the following section, we describe the datasets and the test car
that is used across the experiments. After that, each section describes
an experiment mentioned above.

\subsection{Datasets, test car, and software framework\label{subsec:Experimental-platform}}

\paragraph{PVDN dataset:}

For the evaluation of the object detector performance, the detection
times, and run-times, the PVDN dataset \parencite{Saralajew2021}
is used. This dataset contains 59\,746%

grayscale images with a resolution of 1280$\times$960 pixels where
all light artifacts---both direct (e.\,g., headlamps) and indirect
(e.\,g., light reflections on guardrails)---of oncoming vehicles
are annotated via keypoints. The underlying sequences of the images
are recordings of test drives on rural roads with a single oncoming
vehicle or multiple oncoming vehicles. Several images in the dataset
include artificial light sources like street lamps that increase the
difficulty of detecting light artifacts correctly. As the authors
of the dataset argue, the keypoint annotations allow for an objective
annotation by placing the keypoint on the intensity maximum of each
light artifact. Also, from this, an automatic generation of bounding
boxes is possible, which becomes useful because most of the state-of-the-art
object detectors rely on bounding box annotations. Since those bounding
boxes are inferred automatically, it may happen that one bounding
box covers both direct and indirect instances at the same time. This
is why the task of detecting bounding boxes on the dataset is currently
framed only as a binary classification problem, namely whether the
bounding box covers a relevant light artifact caused by an oncoming
vehicle (either direct or indirect) or not.

The images are frames of recorded video sequences so that the temporal
relationships within the images of a sequence are preserved. Each
scene is recorded with 18\,Hz, either with a short exposure (day
cycle, darker images) or long exposure (night cycle, brighter images).
For the experiments in this work, the day cycle data is used as the
shorter exposure results in a stronger contrast between the background
and light artifacts. Within the PVDN dataset, each illumination cycle
is split into a train, a validation, and a test dataset to enable
the development, evaluation, and testing of algorithms. Most importantly,
the sequences of the dataset contain tags that mark the timestamps
where 
\begin{itemize}
\item the oncoming vehicle is first annotated by its light artifacts, 
\item the driver recognized the oncoming vehicle based on its light artifacts
(indirect or direct), 
\item the vehicle is first directly visible, and 
\item the in-production computer vision system first detected the oncoming
vehicle. 
\end{itemize}
Those tags were collected during the annotation process and the test
group study, which was performed when the dataset was recorded.

\paragraph{Distance evaluation data:}

\begin{figure*}
\subfloat[Camera image with LiDAR point cloud overlaid.]{\begin{centering}
\includegraphics[width=0.25\paperwidth]{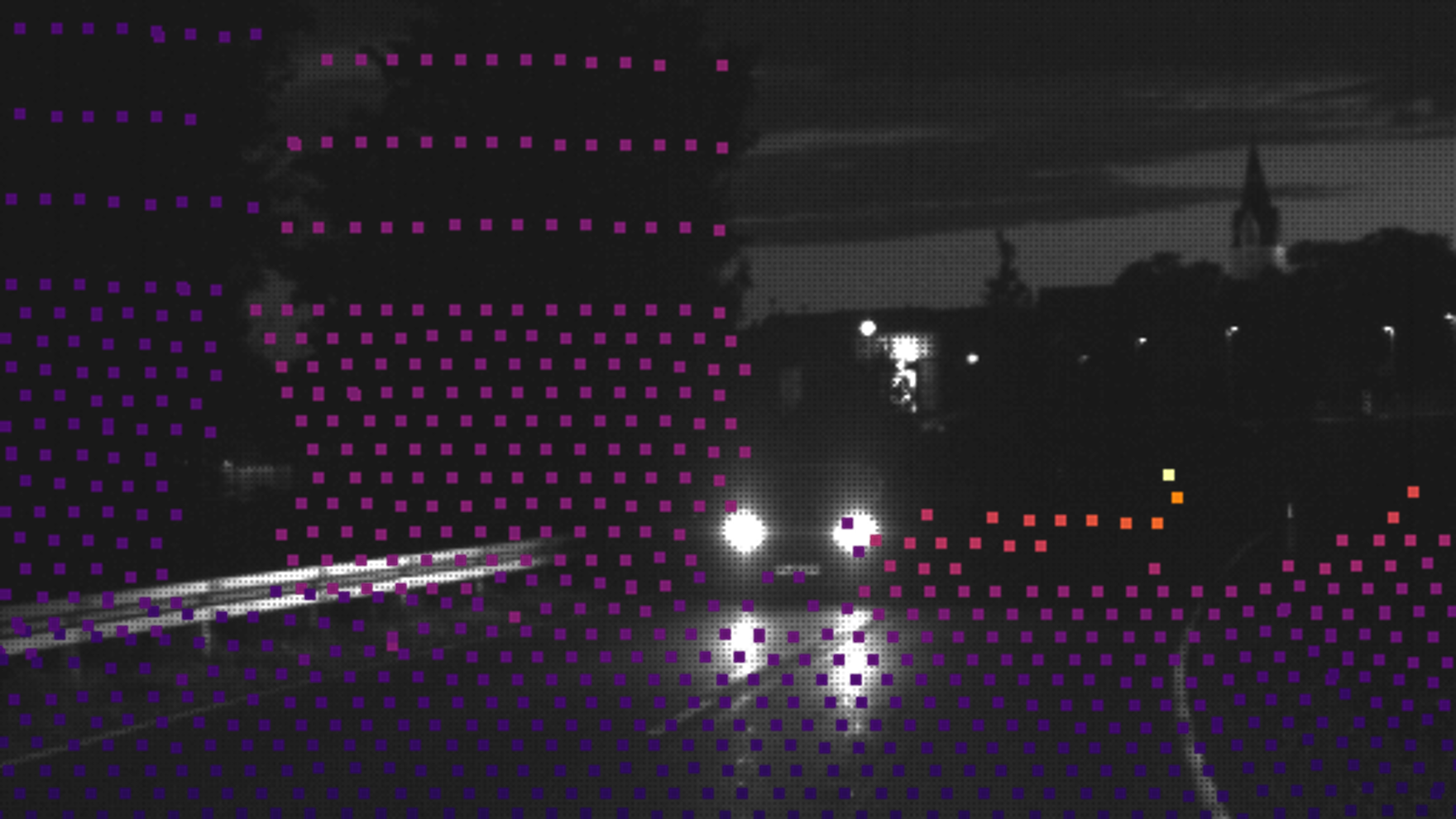}
\par\end{centering}
}\hfill{}\subfloat[Annotated bounding boxes. ]{\begin{centering}
\includegraphics[width=0.25\paperwidth]{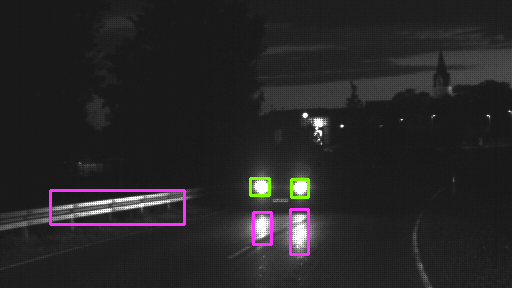}
\par\end{centering}
}\hfill{}\subfloat[Corresponding LiDAR depth image. ]{\begin{centering}
\includegraphics[width=0.25\paperwidth]{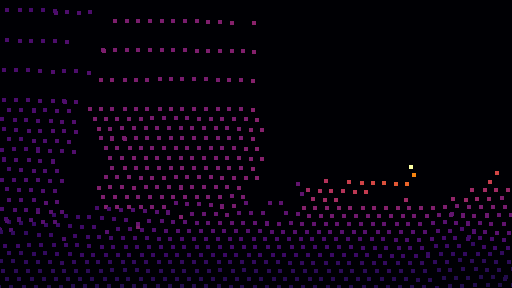}
\par\end{centering}
}

\caption{Exemplary sample of the distance evaluation data. The camera image
is cropped and brightness-adjusted for better visibility. The green
bounding boxes mark direct and the pink indirect light instances.
Black pixels in the LiDAR point cloud mean that there were no measurement
points available at that location. Also, brighter points in the point
cloud indicate greater distances.\label{fig:distance-data-sample}}
\end{figure*}
Since the PVDN dataset does not contain depth data, an additional
small dataset for the evaluation of the distance estimation methods
was recorded. The dataset consists of 24 scenes with 438 images in
total (181 direct and 257 indirect light instances). Each scene contains
five consecutive image frames to later allow for time series analyses.
An exemplary sample is shown in \figref{distance-data-sample}. The
light instances (both direct and indirect ones) were annotated manually
with bounding boxes. The ground truth depth data was captured using
a Hesai LiDAR sensor and the same camera system that was used to record
the PVDN dataset. The single ground truth depth value for each light
instance was calculated using the median of all available depth measurements
within a respective bounding box.

\paragraph{Test car and software framework:}

A test car is used as the platform for deploying the pipeline for
a real use case. It has to be noted that the test car was also used
for recording the PVDN and distance evaluation dataset. Consequently,
the same image input specification as for the PVDN dataset hold. Furthermore,
the test car has a glare-free high beam system \parencite{boke2015mercedes,Knoechelmann2019},
which is used in the experiment for visual demonstrations and for
deploying a provident glare-free high beam system. Each headlight
consists of 84 LEDs, and each is almost illuminating another solid
angle (within a headlight) and can be dimmed independently to all
the other LEDs. Therefore, if an oncoming vehicle is detected and
the glare-free high beam system is activated, the individual LEDs
where the vehicle is located can be turned off such that the overall
headlight system stays in ``high beam mode'' without blinding the
oncoming vehicle---the vehicle moves in a black corridor. This corridor
can be perceived by the driver, making it possible to assess the light
artifact detection quality in a real environment.

To perform the experiments, an additional compute platform is used
consisting of
\begin{itemize}
\item two Intel Xeon CPUs with a base clock frequency of 3.2\,GHz and eight
cores per CPU and
\item one NVIDIA Tesla V100 GPU with 16\,GB RAM\@.
\end{itemize}
\newcommand{\CC}{C\nolinebreak\hspace{-.05em}\raisebox{.4ex}{\tiny\bf +}\nolinebreak\hspace{-.10em}\raisebox{.4ex}{\tiny\bf +}}
\def\CC{{C\nolinebreak[4]\hspace{-.05em}\raisebox{.4ex}{\tiny\bf ++}}}The implementations on this platform are done using Python and \CC\
in the Melodic distribution of the Robot Operating System (ROS).\footnote{\url{https://www.ros.org/}}
The ROS network consists of the following nodes connected sequentially
to each other:
\begin{enumerate}
\item A node that receives the image task-dependent from either the driver
assistance camera or from the dataset;
\item A node that detects the objects according to \subsecref{Object-Detector},
which is internally split into two sub-nodes where one executes the
proposal generation (CPU) and one classifies the proposals by the
shallow NN (GPU);
\item A node that estimates the distances to detected objects according
to \subsecref{Distance-Estimator};
\item A node that performs the tracking according to \subsecref{Object-tracking};
\item A node that publishes the tracked objects task-dependent to either
the test car's CAN bus that controls the headlights or a buffer for
further evaluations.
\end{enumerate}
All nodes run single-threaded on a CPU except the shallow NN, which
is executed on the GPU\@. With running most of the algorithm parts
single-threaded on a CPU, we want to underline the transferability
of the algorithms to a series platform with hardware acceleration.

\subsection{Object detector\label{subsec:Object-detector}}

\begin{table*}
\begin{centering}
\caption{Parameter search space for the optimization of the bounding box annotations.
The context for the specific parameters can be found in \subsecref{Object-Detector}.\label{tab:Parameter-search-space}}
\par\end{centering}
\centering{}%
\begin{tabular*}{1\textwidth}{@{\extracolsep{\fill}}>{\centering}m{0.1\textwidth}>{\raggedright}m{0.4\textwidth}>{\centering}p{0.1\textwidth}>{\centering}p{0.1\textwidth}>{\centering}p{0.1\textwidth}}
\toprule 
Parameter & Description & Search space & Step size & Final value\tabularnewline
\midrule
\midrule 
$\kappa$ & Scaling parameter in dynamic thresholding. & $\left[0.25,0.75\right]$ & 0.05 & 0.4\tabularnewline
\addlinespace
$w$ & Window size in dynamic thresholding. & $\left\{ 5,6,\dots,25\right\} $ & 1 & 19\tabularnewline
\addlinespace
$s$ & Threshold that the mean absolute deviation of a bounding box has to
exceed to be proposed. & $\left[0,0.1\right]$ & 0.01 & 0.01\tabularnewline
\addlinespace
$d$ & Maximal $L_{\infty}$ distance that is allowed between blobs to be
considered in the same bounding box. & $\left\{ 1,2,\dots,9\right\} $ & 1 & 4\tabularnewline
\bottomrule
\end{tabular*}
\end{table*}
The object detector described in \subsecref{Object-Detector} is conceptually
equivalent to the detector proposed by \textcite{Saralajew2021} and
consists of a rule-based proposal generation algorithm and an NN-based
classifier. Originally, the parameters of the proposal generation
algorithm were selected by a random search using the PVDN dataset.
To improve this selection, a hyperparameter search for the proposal
generation algorithm is performed using the tree-structured Parzen
estimator approach \parencite{bergstra2011}. This approach belongs
to the family of sequential model-based optimization approaches and
is a standard algorithm for hyperparameter optimization. Before defining
the objective function, the bounding box quality score and the events
to define the F-score, recall, and precision is introduced \parencite{Saralajew2021}:
The goal of the bounding box quality score is to define a measure
to assess the quality of a bounding box prediction algorithm by using
ground truth keypoints. Because each light artifact is annotated by
exactly one keypoint and each bounding box should span exactly one
light artifact, in the best case,
\begin{itemize}
\item each ground truth keypoint lies within exactly one predicted bounding
box, and 
\item each predicted bounding box spans around exactly one ground truth
keypoint.
\end{itemize}
To formalize this concept, the following events for keypoints and
bounding boxes are introduced:
\begin{itemize}
\item True positive: The ground truth keypoint is covered by at least one
bounding box (light artifact covered);
\item False negative: The ground truth keypoint is not covered by any bounding
box (light artifact not covered);
\item False positive: The bounding box does not cover any ground truth keypoint
(no light artifact covered).
\end{itemize}
By using these events, the F-score, precision, and recall can be computed.
Additionally, to quantify the quality of true-positive bounding boxes,
the following quantities are calculated:
\begin{itemize}
\item $n_{K}\left(b\right)$: The number of ground truth keypoints in the
true-positive bounding box $b$;
\item $n_{B}\left(k\right)$: The number of true-positive bounding boxes
that cover the keypoint $k$.
\end{itemize}
To convert these numbers into values in the range $\left[0,1\right]$,
where one means best performance and zero worst, the reciprocal value
is taken. Finally, the values are averaged across the dataset to obtain
measures for the performance of a detector:
\begin{align}
q_{K} & =\frac{1}{N_{B}}\sum_{b}\frac{1}{n_{K}\left(b\right)},\\
q_{B} & =\frac{1}{N_{K}}\sum_{k}\frac{1}{n_{B}\left(k\right)},
\end{align}
where $N_{B}$ is the total number of true-positive bounding boxes,
and $N_{K}$ is the total number of keypoints covered by bounding
boxes. These two measures quantify the uniqueness of predicted bounding
boxes that are true positives with respect to how many keypoints are
contained within a bounding box and how many bounding boxes cover
the same keypoint. For example, in an image with several keypoints,
$q_{K}$ is low if a large bounding box spans over the whole image.
The overall bounding box quality is determined by $q=q_{K}\cdot q_{B}$,
where a value of one indicates best performance and zero worst.

Using the definition of the bounding box quality $q$ based on the
introduced events, the objective function $h\left(\theta\right)$
of the hyperparameter optimization is 
\begin{equation}
h\left(\theta\right)=1-q\left(\theta\right)\longrightarrow\min,
\end{equation}
where $\theta$ is a hyperparameter configuration. This objective
function encourages bounding box generators (proposal generation algorithms)
where each keypoint is uniquely assigned a bounding box. The specific
search space configuration can be found in \tabref{Parameter-search-space}.

\begin{table*}
\caption{Performance results of the proposed \emph{optimized} detector, the
\emph{baseline} detector of \textcite{Saralajew2021}, and \emph{YOLOv5}
architectures (trained on the optimized bounding box annotations)
on the PVDN day-test dataset. The values in parentheses represent
the performance values of the generated bounding box annotations on
the PVDN day-test dataset. These performance values for the annotations
represent the maximum performance that can be achieved with the proposal
generation algorithm with respect to the performance metrics. For
the baseline model, the performances are reported for two different
image sizes according to the published results in the GitHub repository.
The run-times are measured on the specified platform without the ROS
framework. For the baseline and the optimized model, the run-times
are reported with standard deviations as the run-time for a specific
image depends on the number of region proposals. Also, it has to be
noted that the proposal generation part of the baseline and optimized
model are always executed on a CPU\@. Acronyms: \textquotedblleft Par.\textquotedblright{}
is the number of parameters; \textquotedblleft Prec.\textquotedblright{}
is the precision.\label{tab:object-detector-results}}

\centering{}%
\begin{tabular*}{1\textwidth}{@{\extracolsep{\fill}}cccccccccc}
\toprule 
Model & Image size & Par. {[}M{]} & Run-time {[}ms{]} & Prec. & Recall & F-score & \emph{$q$} & $q_{K}$ & $q_{B}$\tabularnewline
\midrule
\midrule 
\multirow{4}{*}{Baseline} & \multirow{2}{*}{345$\thinspace\times\thinspace$240} & 0.9 & 18.2$\thinspace\pm\thinspace$3.3 & 0.88 & 0.54 & 0.67 & 0.40 & 0.40$\thinspace\pm\thinspace$0.21 & 1.00$\thinspace\pm\thinspace$0.00\tabularnewline
 &  & -- & -- & (1.00) & (0.69) & (0.81) & (0.42) & (0.42$\thinspace\pm\thinspace$0.24) & (1.00$\thinspace\pm\thinspace$0.00)\tabularnewline
 & \multirow{2}{*}{640$\thinspace\times\thinspace$480} & 0.9 & 52.71$\thinspace\pm\thinspace$2.38 & 0.90 & 0.64 & 0.75 & 0.48 & 0.48$\thinspace\pm\thinspace$0.26 & 1.00$\thinspace\pm\thinspace$0.00\tabularnewline
 &  & -- & -- & (1.00) & (0.72) & (0.84) & (0.50) & (0.50$\thinspace\pm\thinspace$0.28) & (1.00$\thinspace\pm\thinspace$0.02)\tabularnewline
\addlinespace
\multirow{2}{*}{Optimized} & \multirow{2}{*}{640$\thinspace\times\thinspace$480} & 0.9 & 21.98$\thinspace\pm\thinspace$1.63 & 0.85 & 0.80 & 0.82 & 0.69 & 0.69$\thinspace\pm\thinspace$0.30 & 1.00$\thinspace\pm\thinspace$0.02\tabularnewline
 &  & -- & -- & (1.00) & (0.87) & (0.93) & (0.70) & (0.70$\thinspace\pm\thinspace$0.30) & (1.00$\thinspace\pm\thinspace$0.00)\tabularnewline
\addlinespace
YOLOv5s & \multirow{3}{*}{960$\thinspace\times\thinspace$960} & 7.0 & 14.8 & 0.98 & 0.67 & 0.80 & 0.67 & 0.70$\thinspace\pm\thinspace$0.31 & 0.97$\thinspace\pm\thinspace$0.12\tabularnewline
YOLOv5x &  & 86.1 & 28.1 & 0.99 & 0.76 & 0.86 & 0.67 & 0.69$\thinspace\pm\thinspace$0.30 & 0.98$\thinspace\pm\thinspace$0.10\tabularnewline
 &  & -- & -- & (1.00) & (0.87) & (0.93) & (0.70) & (0.70$\thinspace\pm\thinspace$0.30) & (1.00$\thinspace\pm\thinspace$0.00)\tabularnewline
\bottomrule
\end{tabular*}
\end{table*}
We optimized the hyperparameters on the official PVDN training set,
selected the best parameters based on the objective function value
on the validation set, and reported the results on the test set. These
optimized hyperparameters were then used to generate \emph{an optimized
set of bounding box annotations for the PVDN dataset:} bounding boxes
that cover a keypoint (a light artifact) were kept as ground truth
bounding boxes. With these optimized bounding box annotations, the
classifier was trained to distinguish between bounding boxes that
contain light artifacts and bounding boxes that do not. Similar to
the hyperparameter optimization, all three dataset splits of the PVDN
dataset were used accordingly to train the proposal classifier. The
classifier was trained for 300 epochs with an initial learning rate
of 0.001, batch size of 64, weight decay of 0.01, and binary cross-entropy.
Moreover, the Adam optimizer \parencite{kingma2014} was used, and
images were augmented with horizontal flips, rotations, crops, and
gamma corrections while training. The confidence threshold for a valid
classification of a light artifact was set to 0.5. To foster public
use and to ensure reproduction, the whole pipeline implemented in
Python with the deep learning framework PyTorch\footnote{\url{https://pytorch.org/}}
is publicly available.\footnote{\url{https://github.com/larsOhne/pvdn}}
In order to make our custom detection approach comparable to state-of-the-art
end-to-end object detection algorithms, we also report the performance
of both YoloV5s and YoloV5x.

\tabref{Parameter-search-space} shows the results of the hyperparameter
optimization. Using these parameters to generate the bounding box
annotations, the optimized proposal generator achieves the results
reported in \tabref{object-detector-results} (see values in parentheses).
The optimized proposal generation algorithm increases the bounding
box quality $q$ from 50\,\% to 70\,\% and increases the F-score
from 84\,\% to 93\,\%. Therefore, the optimized proposal generation
algorithm shows a clear improvement of the automatically inferred
bounding box annotations compared to the baseline. However, the performance
of the optimized region proposal algorithm is still not optimal because
not each performance value is 100\,\%. The scores which are at 100\,\%
must be at this level due to the construction principle of the bounding
boxes: Each ground truth bounding box in the generated annotations
is a valid bounding box so that the precision must be 100\,\%; It
is likely that each ground truth bounding box only covers one keypoint
due to the construction of the bounding boxes by non-maximum suppression
so that the quality $q_{B}$ must be close to 100\,\%. 

Because of the improved performance of the optimized proposal generation
algorithm, the trained object detector shows a significant improvement
of the detection performance too, as simply more of the light artifacts
are captured by the proposal generation algorithm---for instance,
notice the improvement of the F-score by 7\,\% and of the bounding
box quality $q$ by 21\,\%. Nevertheless, the performance values
suggest that the optimized object detector can be further improved
because, for example, there is a difference of 11\,\% between the
achieved F-score and the achievable with respect to the generated
optimized bounding box annotations. Considering the F-score of the
two YOLOv5 variants, the optimized custom object detector is slightly
superior to YOLOv5s but somewhat inferior to YOLOv5x. Moreover, the
optimized object detector is marginally better with respect to the
bounding box quality (achieves almost the best possible bounding box
quality). The run-times of the models clearly show that YOLOv5s is
the fastest detector. However, it must be noted that the optimized
detector is just approximately 7\,ms slower and not as optimized
as the YOLOv5s architecture. The proposal generation of the optimized
detector is single-threaded executed on a CPU, which is the reason
why, on average, 20.55\,ms of the overall run-time are consumed by
this step. Therefore, it must be expected that the run-time can be
greatly improved by optimizing the execution of the proposal generation
step.

In summary, for embedded and safety-critical purposes, our custom
approach is to be chosen before YOLOv5. Because it has significantly
fewer parameters and thus requires less memory, and has high optimization
potentials for embedded hardware. Also, the proposed optimized object
detector consisting of a rule-based proposal generation algorithm
and a shallow NN-classifier is more transparent than the deep NN-architectures
of YOLOv5. Therefore, the approach is easier to validate and verify
for safety-critical applications since the model behavior can be better
understood and interpreted. Finally, we conclude that the optimized
and trained light artifact detector sets a new benchmark for the PVDN
dataset.

\subsection{Distance estimation\label{subsec:Distance-estimation}}

\begin{figure}
\begin{centering}
\includegraphics[viewport=8bp 2bp 245bp 163bp,clip,width=0.95\columnwidth]{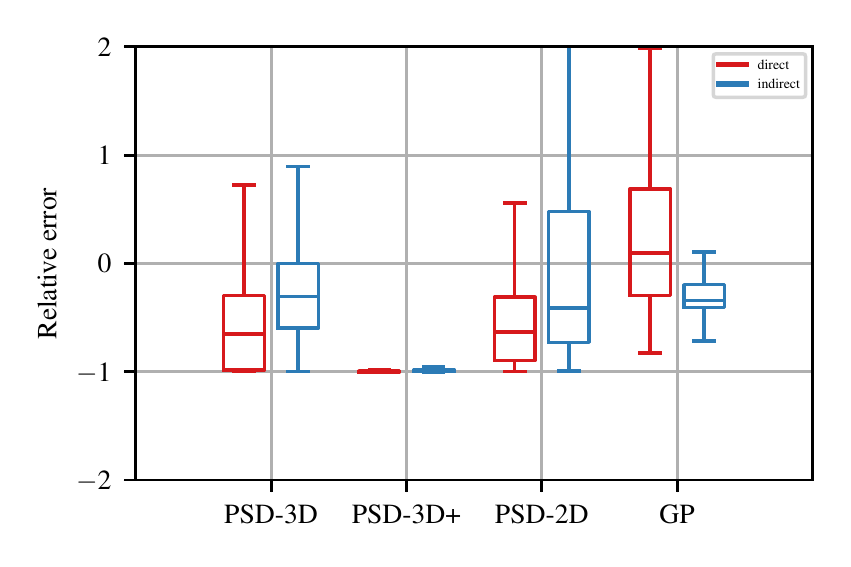}
\par\end{centering}
\caption{Relative error results of the different methods applied on each single
light artifact.\label{fig:perf-single-img}}
\end{figure}

This section presents the evaluation of the proposed methods of \subsecref{Distance-Estimator}
on the dataset described in \subsecref{Experimental-platform}. First,
the general performance of each method is evaluated on the available
data. For that, each light artifact marked by a bounding box is transformed
to a \emph{single pixel by taking the center of the bounding box.}
Performance results are shown in \figref{perf-single-img}. It becomes
clear that with a median relative error of $0.12$ for direct and
$-0.32$ for indirect light instances, the Ground Plane approach (GP),
which only considers the area ahead as a plane, outperforms approaches
PSD-3D, PSD-3D\nolinebreak[4]\hspace{-.05em}\raisebox{.4ex}{\tiny\bf +},
and PSD-2D, which try to estimate the road geometry using the PSD\@.
A negative error means that the estimated distance is less than the
ground truth distance. An in-depth analysis of the PSD shows that
the positioning of the vehicle on the road described by the PSD is
often too inaccurate to give a precise enough representation of the
exact road geometry ahead, which is needed in order for the PSD approaches
to work. Especially in curves, an accurate positioning of the vehicle
on the road segment is absolutely mandatory since even a slight deviation
can cause a considerable discrepancy between the actual road geometry
ahead and the one described by the PSD at a specific time step. 

When looking at the performance of the GP method, a performance deficit
between direct and indirect light instances becomes clear. There are
several possible reasons for this. First, the direct light instances
are always located further away from the ego-vehicle than the indirect
ones. The direct instances in the underlying data have an average
distance of 83\,m, whereas indirect instances are, on average, 63\,m
away. Therefore, inaccuracies influence the relative error more for
the indirect instances. Second, indirect light instances often span
over a large area (e.\,g., on the street), where the acquisition
of a single ground truth distance value is difficult as the beginning
of the annotated area has a different distance value than the end.
This can lead to partly inaccurate ground truth values. Third, all
of the mentioned methods strongly depend on the quality of the intrinsic
and extrinsic camera calibration. Thus, unknown inaccuracies in the
calibration can also affect the result. Finally, the assumption of
the environment ahead being a plane could often be inaccurate. If
the assumption were true, the expected result for indirect light instances
on the road would be nearly perfect compared to the ground truth,
whereas all light instances located above the road (e.\,g., headlights
or light reflections on guardrails) would give a too far away distance,
as the camera ray would intersect with the plane behind the actual
light artifact. However, the results show that the direct light instances
are nearly perfect (only a little overshooting distance estimation
of approximately 12\,\%), whereas the distance estimation for indirect
instances falls too short. This indicates that the data often contains
scenes where the ground plane assumption does not hold.

\begin{figure}
\begin{centering}
\includegraphics[viewport=0bp 2bp 232bp 157bp,clip,width=0.95\columnwidth]{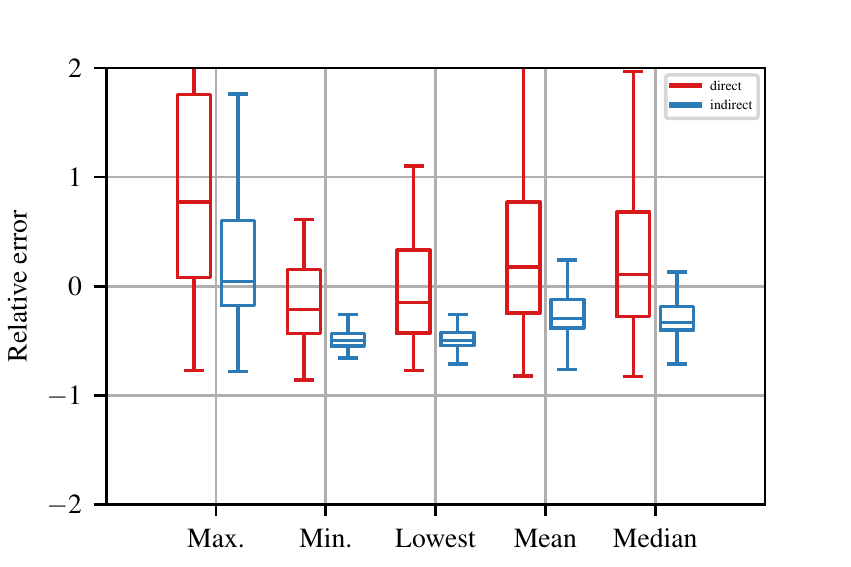}
\par\end{centering}
\caption{Relative error results of the GP method by taking the depth values
of all pixels within a bounding box into account.\label{fig:perf-bbox}}
\end{figure}

To investigate whether the distance estimation can be improved by
considering the distance estimate of each pixel within a bounding
box, several heuristics are analyzed. As the methods using the PSD
already did not show satisfying results in the first experiment, the
following evaluations are only done for the GP method. To retrieve
the final distance estimation from all distance values within a bounding
box, five simple approaches were compared with each other:
\begin{enumerate}
\item only considering the maximum distance value;
\item only considering the minimum distance value;
\item only considering the distance value of the lowest pixel in the bounding
box (as it is closest to the estimated plane); 
\item taking the mean over all distance values in the bounding box;
\item taking the median over all distance values in the bounding box.
\end{enumerate}
The results are shown in \figref{perf-bbox}. Interestingly, the five
approaches do hardly show any improvements. Only the approach of taking
the maximum distance value within a bounding box improves the distance
estimation for indirect light instances, which makes sense considering
that the original estimation for indirect light instances was often
too short.

\begin{figure}
\centering{}\includegraphics[viewport=8bp 8bp 240bp 163bp,clip,width=0.95\columnwidth]{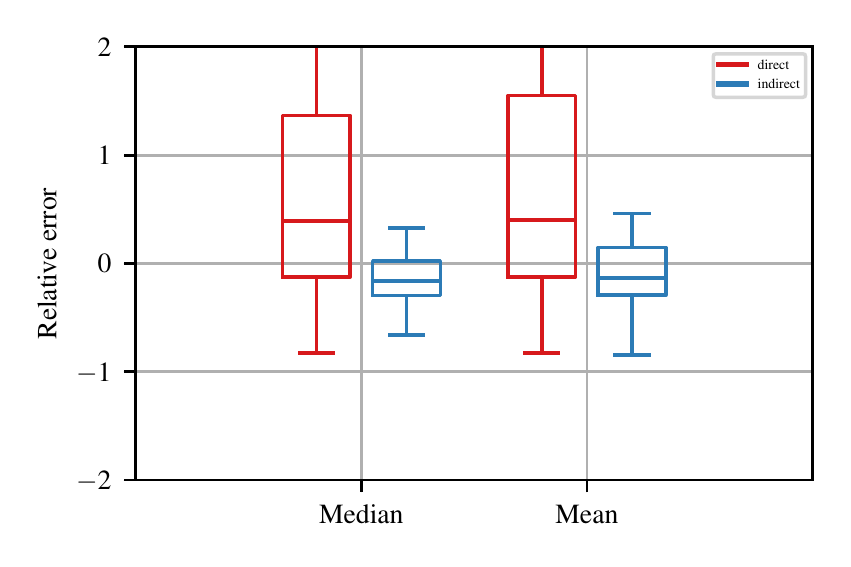}\caption{Relative error results of taking the mean and the median of a series
of five consecutive light artifacts.\label{fig:perf-timeseries}}
\end{figure}

Since the annotation format of the dataset was chosen so that the
correspondence of light instances across multiple frames can be determined,
in a final experiment, the distance estimation for the GP method is
attempted to be stabilized by considering a series of consecutive
distance estimations for the same instance. The idea is that with
this, possible outliers can be filtered. For that, the two approaches
of taking the median or the mean of a series of distance estimations
are compared. One series consists of five consecutive images. The
results can be seen in \figref{perf-timeseries} and do not show a
significant improvement or stabilization of the distance estimations.
The relative errors show an offset by roughly the same positive amount,
which is a reasonable behavior since the predictions from previous
time steps, where the instances were still further away, increase
the final estimated distance. Note that this approach requires a tracking
of detected objects across multiple images.

To summarize, the high positioning inaccuracy of the ego-vehicle on
the road described by the PSD results in a highly inaccurate distance
estimation of light artifacts. However, the core idea itself is promising,
as it models the road environment ahead in its actual shape. Yet,
the current positioning inaccuracies make the data unusable for this
task. This limitation could be tackled by using high-definition maps
and visual odometry to improve localization in the future. The alternative
approach of modeling the world ahead as a simple plane and finding
the intersection with the camera ray, however, shows satisfying results.
Another advantage is that this approach does not require any sensor
data except for the camera input and its calibration and also comes
at a very low computational cost since calculating the intersection
of a line with a plane requires only a few floating-point operations.
Still, when an object is not located directly on the ground (e.\,g.,
light reflections on guardrails), the method becomes inaccurate, too.
For future improvements, approaches should be analyzed that try to
estimate the surface of the environment ahead in order to account
for curvatures of the road surface and thus return a better approximation
than the simple plane \parencite[e.\,g.,][]{P.F.Alcantarilla.2011}.
However, for the system presented in this paper, the accuracy of this
method is considered to be sufficient, and, therefore, the GP approach
is used as the distance estimation module in our system.

\subsection{Time benefit\label{subsec:Time-benefit}}

The goal of this experiment is to evaluate the time benefit of the
proposed system in terms of a provident detection of oncoming vehicles
with respect to human performance and an in-production computer vision
system for vehicle detection at night. For this purpose, evaluations
are performed on the test and validation dataset (to increase the
database) of the PVDN dataset (see \subsecref{Experimental-platform}).
The sequences from the dataset are processed by the implemented ROS
network, and the detection times are computed on the image frame level
of the dataset. This means that the number of frames is counted between
the first indirect sight annotation is available (object detectable
by human annotator) and the respective system recognizing the object
in the image. By knowing that the frame rate of the camera is 18\,Hz,
the counted number of frames can be converted into seconds. Note that
these computed times are without the system-specific processing time
(see \ref{subsec:Provident-glare-free-high}) and that the images
in the dataset include a tag that specifies when the in-production
system detected the oncoming vehicle. Additionally, note that even
if the dataset contains images with multiple oncoming vehicles, only
one can become visible first because the images are recorded on a
two-lane road.\footnote{A road with one lane in each direction.}
Therefore, the number of frames is counted with respect to this first
vehicle by using the available keypoint semantics (keypoints in the
dataset are associated with the vehicles). In this context, the frame
number when the single frame or tracking-based detection recognizes
an oncoming vehicle is determined by the first bounding box that includes
a keypoint associated with the first vehicle. 

\begin{figure}
\begin{centering}
\includegraphics[width=0.95\columnwidth]{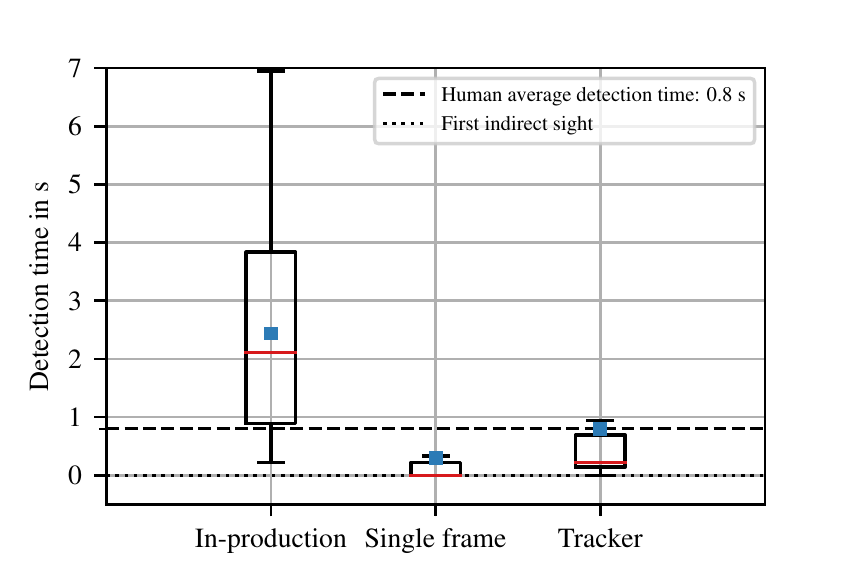}
\par\end{centering}
\caption{Detection times of the oncoming vehicle after the first indirect sight
of the in-production computer vision system, the proposed system based
on single frame detections (without tracker), and with object tracking
(plausibility checker). Additionally, the human detection performance
is presented as a constant average value of 0.8\,s after the first
perceivable light artifact of the oncoming vehicle \parencite{Oldenziel2020}.\label{fig:detection_times}}
\end{figure}
\figref{detection_times} shows the results in the form of a box plot
on the respective 39 sequences of the PVDN dataset. In 18 sequences,
the in-production computer vision system did not detect a vehicle
and thus has fewer measurement samples. It can be seen that the proposed
system based on the tracker detects oncoming vehicles, on average,
1.6\,s faster than the in-production computer vision system and is
as fast as a human on average. The first detection based on a single
frame is, on average, 2.1\,s before the detection of the in-production
computer vision system and 0.5\,s before the detection of a human.
The delay between the single frame detection and the tracker is caused
by the plausibility phase of the tracker: an object has to be detected
for at least five frames before it is sent as output. Five frames
correspond to approximately 277\,ms, and this is the minimal delay
inherently caused by the plausibility checker (compare with \figref{motivation}).
Therefore, it is not surprising that the time difference between the
single frame detection and the tracker is, on average, 500\,ms. However,
overall, the results clearly show the considerable time benefit that
can be achieved by such a sensing system.

\subsection{Provident glare-free high beam\label{subsec:Provident-glare-free-high}}

To demonstrate the usefulness of the provident detection information
for ADAS functionalities, we integrated the proposed detection system
into the test car and used the provident detection information to
control the adaptive headlights. By doing so, a provident glare-free
high beam functionality is realized. The results of this experiment
provide useful information about the applicability in real use cases: 
\begin{itemize}
\item it shows that the entire workflow of the detection system can run
in real time; 
\item it nicely visualizes the detection results in a real environment;
\item it shows that glare-free high beam functions can be implemented without
blinding oncoming vehicles due to latencies in the computer vision
system (see \figref{motivation} and discussion in \subsecref{System-related-latencies}).
\end{itemize}
The glare-free high beam functionality is suitable to visualize the
detection results, as the headlights can be considered as projectors
that visualize the detected objects by turning off the respective
pixels. Therefore, any serious inaccuracy in the system becomes immediately
visible, and thus the integration serves as a proof of concept whether
the object localization uncertainties are in such a range that they
still provide useful information for later systems. As already said,
for this experiment, the proposed detection pipeline is integrated
into the test car, and test drives are performed on \emph{public}
rural roads at night. To ensure that other drivers are not put at
risk, the light artifact detection pipeline is integrated in such
a way that the detection results of our pipeline are muted as soon
as the in-production system detects a vehicle. Thus, the integration
of the proposed pipeline just bridges the time difference between
the provident detection and the in-production system. To have a unique
light artifact detection output, only the detected object (after tracking)
with the highest intensity value is sent to the glare-free high beam
module. Because the system is tested on two-lane roads, this ensures
that the detected light artifact with the highest intensity \emph{always
converges} to the vehicle's headlamps since, after direct sight to
the vehicle, the intensity maximum is in a bounding box of the headlamps.
This concept circumvented the need 
\begin{itemize}
\item to classify light artifacts into direct and indirect light instance, 
\item to cluster direct light instances to vehicle bounding boxes,
\item to associated light artifacts to vehicles, and
\item to detect occurrence points of vehicles because the proposed detection
pipeline locates light artifacts.
\end{itemize}
\begin{figure}
\begin{centering}
\includegraphics[width=0.95\columnwidth]{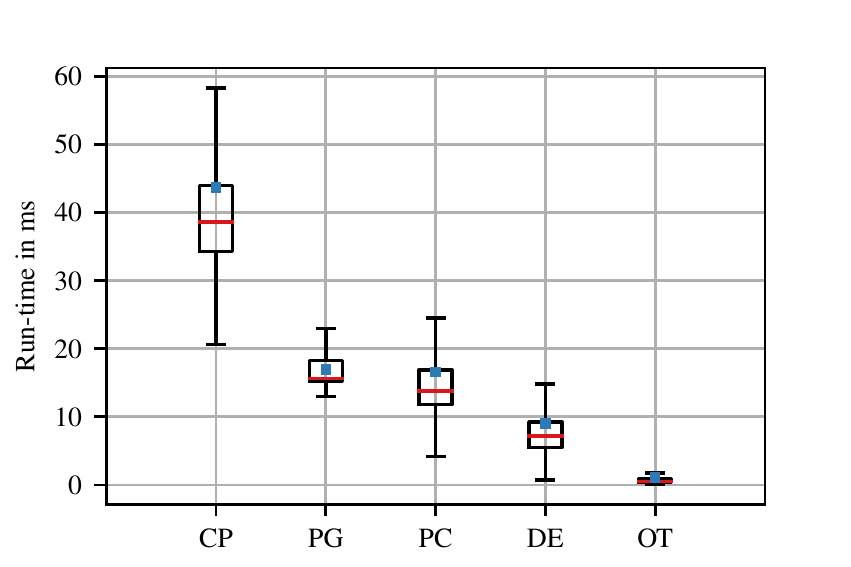}
\par\end{centering}
\caption{Run-times of the modules (described from left to right): Run-times
of the Complete Pipeline (CP), Proposal Generation (PG), Proposal
Classification (PC), Distance Estimator (DE), and Object Tracker (OT).\label{fig:Run-times}}
\end{figure}
Before the system is tested on public roads, the real-time capabilities
of the pipeline are analyzed by measuring the computation times on
the test car's hardware (see \subsecref{Experimental-platform}).
In this context, the run-times are determined by the elapsed time
(ROS time) from receiving the node input to publishing the output.
For instance, the run-time of the proposal classifier is the time
from receiving the input in the form of the bounding boxes and the
image until the classification of all bounding boxes is determined
and published. Since the camera captures images with 18\,Hz, the
requirement is that the entire pipeline has an execution time faster
than 18\,Hz. 

\figref{Run-times} presents the run-time analysis in the form of
a box plot. The measurements were performed on the 7\,030 images
(test and validation dataset) of the PVDN dataset. The average run-time
of the complete pipeline (entire ROS network) for one image is on
average 0.044\,s so that the real-time requirement is fulfilled.
However, it must be noted that the run-time of the pipeline is not
constant. For example, the run-time is strongly affected by the number
of bounding boxes created by the proposal generator. Moreover, with
an increasing number of components (after the dynamic thresholding
step) during the proposal generation, the run-time of the bounding
box creation (inside the proposal generator) increases as well. Overall,
the real-time requirement is fulfilled for 90\,\% of the analyzed
images (see \figref{Run-times}), and, therefore, the system can be
deployed in the test car.\footnote{If the computation is still running and the camera has already captured
a new image, the new camera image is dropped in the deployed algorithm.} It must be noted that the run-times of the proposal classifier and
the proposal generator are different from the run-times reported in
\subsecref{Object-detector}. This is because the evaluation here
is conducted on the test and validation dataset and that the ROS framework
causes a non-neglectable overhead due to data transformations to publish
and process messages.

\begin{figure*}
\begin{centering}
\subfloat[First sight of indirect light instances on the guardrail between the
trees.\label{fig:Visual-demonstration-start}]{\centering{}\includegraphics[viewport=0bp 0bp 443.674bp 263.5987bp,clip,width=0.8\columnwidth]{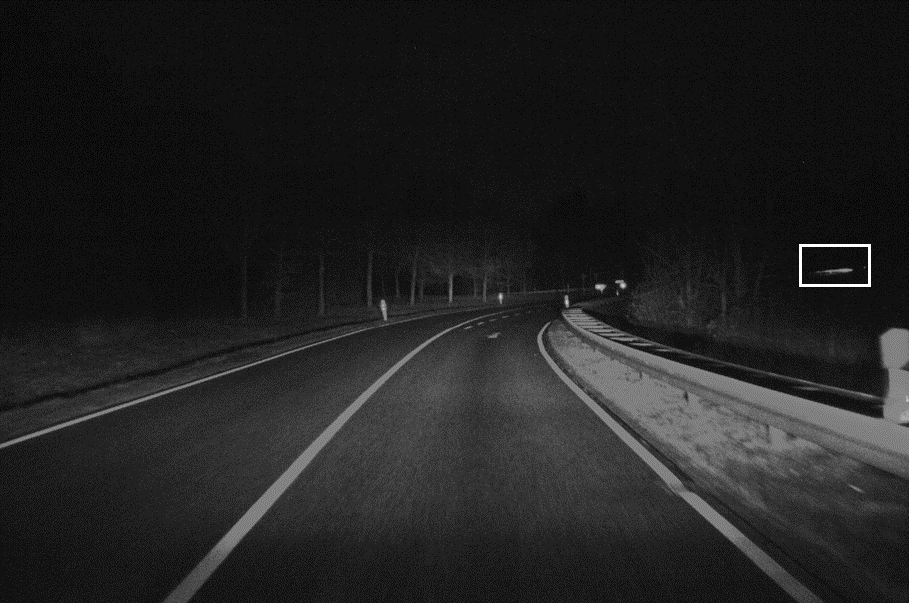}\hspace{4em}\includegraphics[viewport=0bp 0bp 444.163bp 263.5987bp,clip,width=0.8\columnwidth]{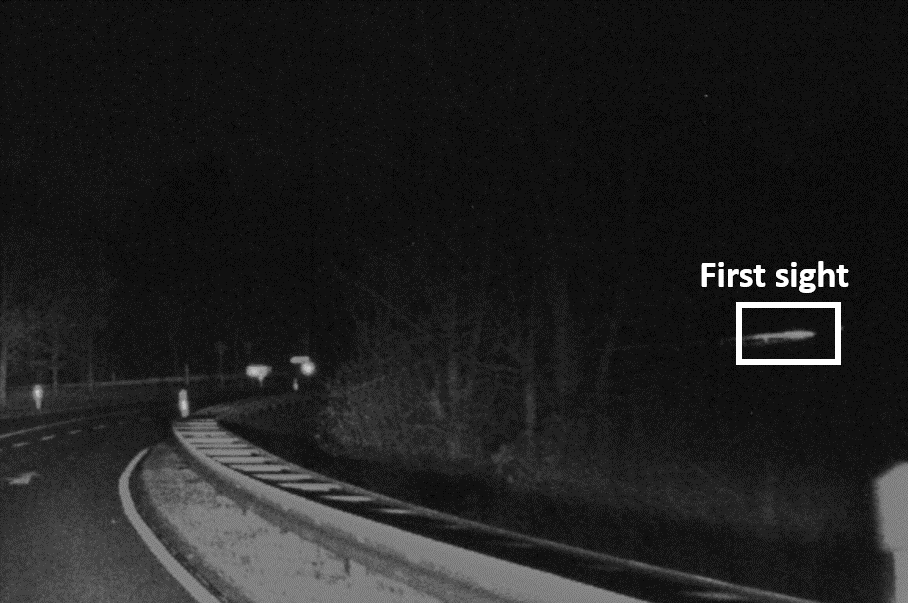}}
\par\end{centering}
\begin{centering}
\subfloat[First detection of the system based on single frame detections 0.5\,s
after first sight. The detection contains two direct instances (headlamps)
as well as an indirect instance on the guardrail. Note that the entire
scene is still illuminated by the adaptive headlight (see the bright
lane markings and the trees).\label{fig:Visual-demonstration-single-frame}]{\centering{}\includegraphics[viewport=0bp 0bp 443.674bp 263.6364bp,clip,width=0.8\columnwidth]{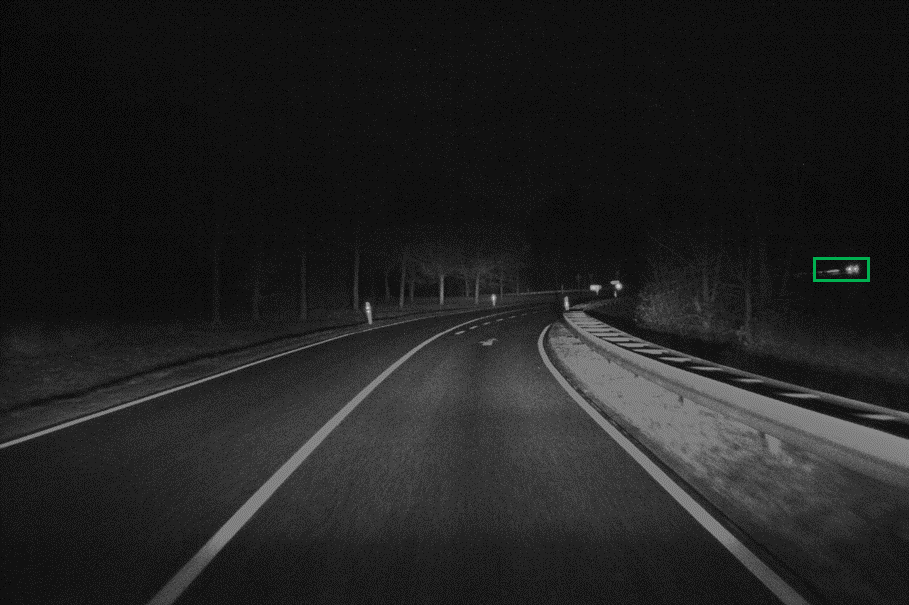}\hspace{4em}\includegraphics[viewport=0bp 0bp 444.203bp 263.6364bp,clip,width=0.8\columnwidth]{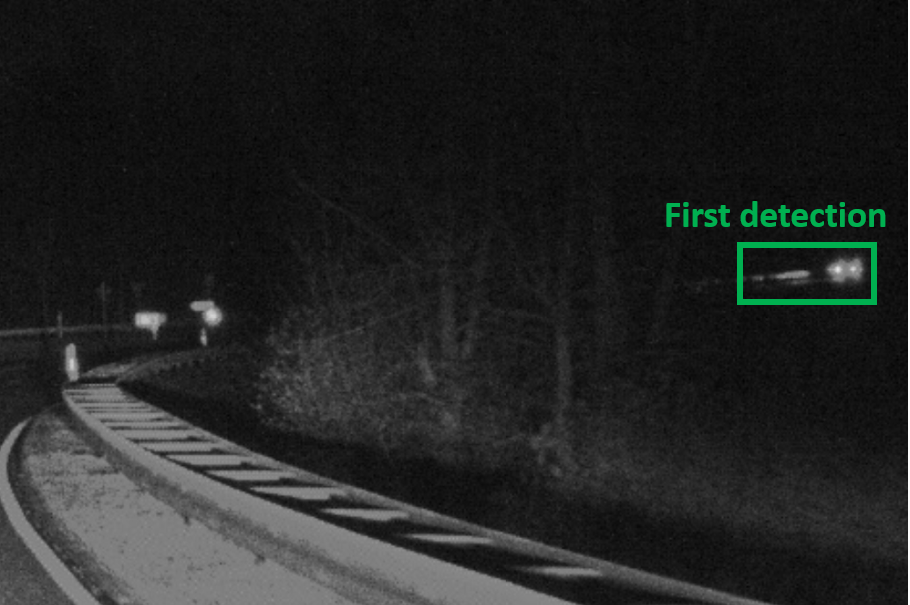}}
\par\end{centering}
\begin{centering}
\subfloat[First detection of the tracker 2.6\,s after first sight and, the
headlights dim the corresponding pixels. The system's reaction is
based on the indirect light instance on the guardrail as the vehicle
itself is not visible in this frame. The dimmed region is best recognized
by focusing on the glow changes of the lane markings and that trees
are not illuminated. Also, note how the indirect light instance with
the highest intensity is at the position where the vehicle will occur,
which shows how this light artifact converges to the vehicle position.\label{fig:Visual-demonstration-tracker}]{\centering{}\includegraphics[viewport=0bp 0bp 444.163bp 263.5987bp,clip,width=0.8\columnwidth]{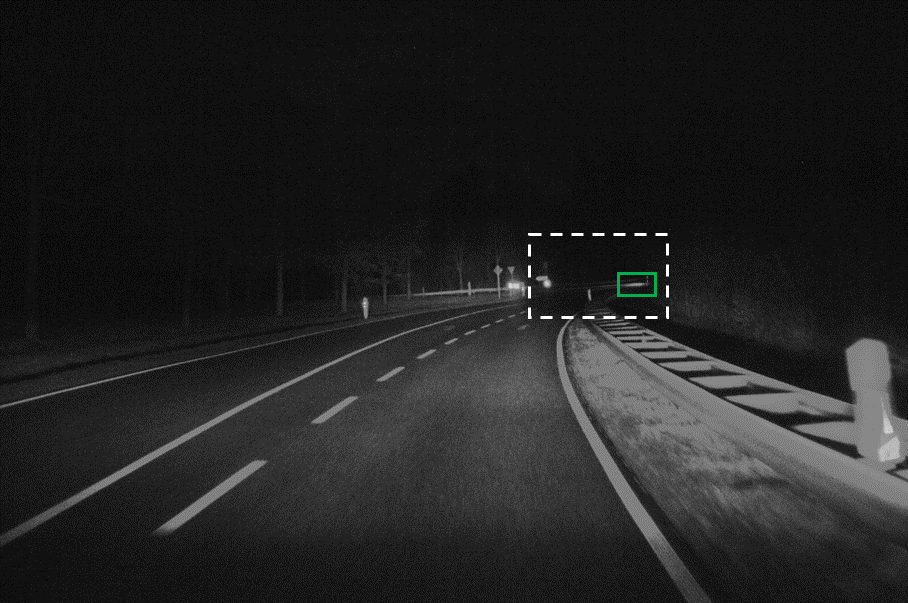}\hspace{4em}\includegraphics[viewport=0bp 0bp 443.591bp 263.5987bp,clip,width=0.8\columnwidth]{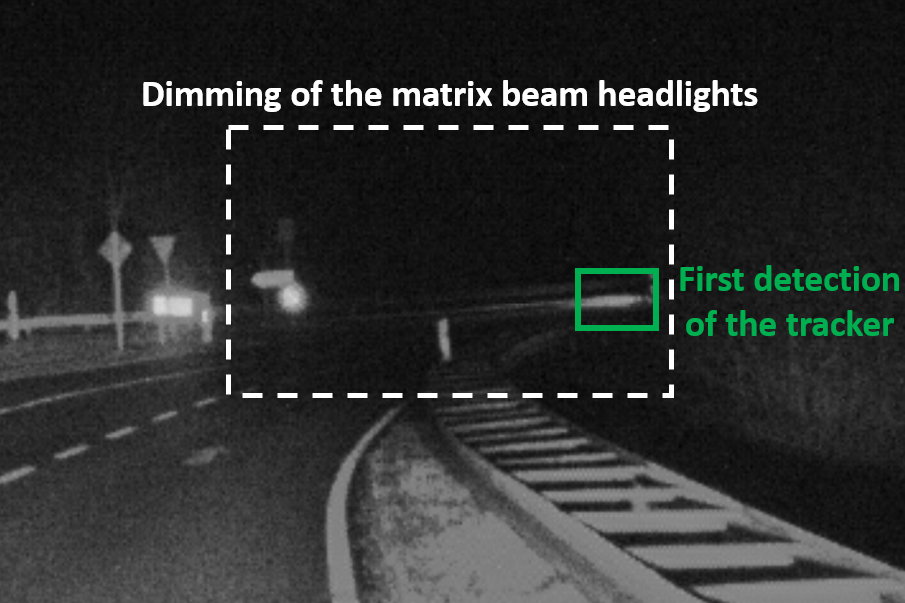}}
\par\end{centering}
\begin{centering}
\subfloat[First detection of the in-production system based on direct light
instances 3.8\,s after first sight.\label{fig:Visual-demonstration-adas}]{\centering{}\includegraphics[viewport=0bp 0bp 444.163bp 264.0728bp,clip,width=0.8\columnwidth]{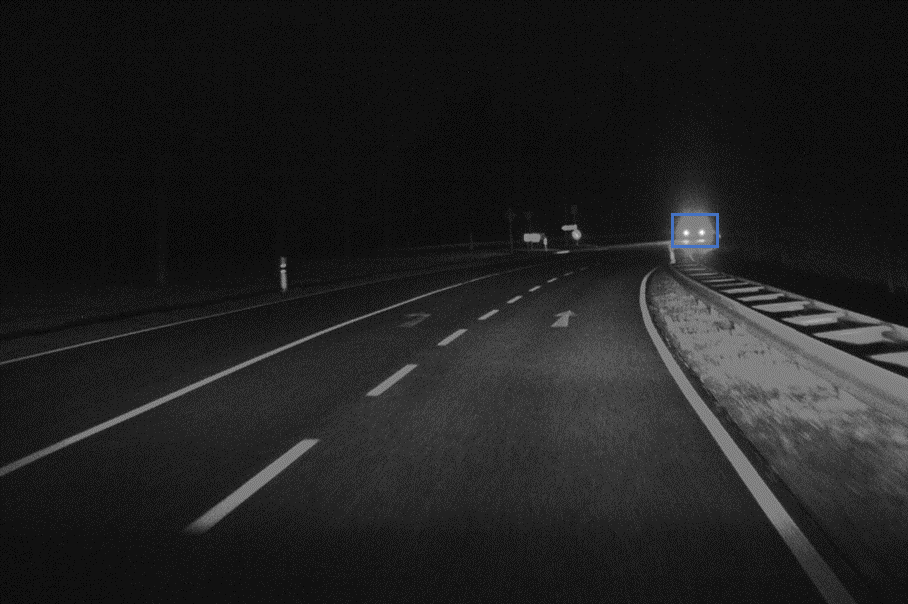}\hspace{4em}\includegraphics[viewport=0bp 0bp 444.163bp 264.1089bp,clip,width=0.8\columnwidth]{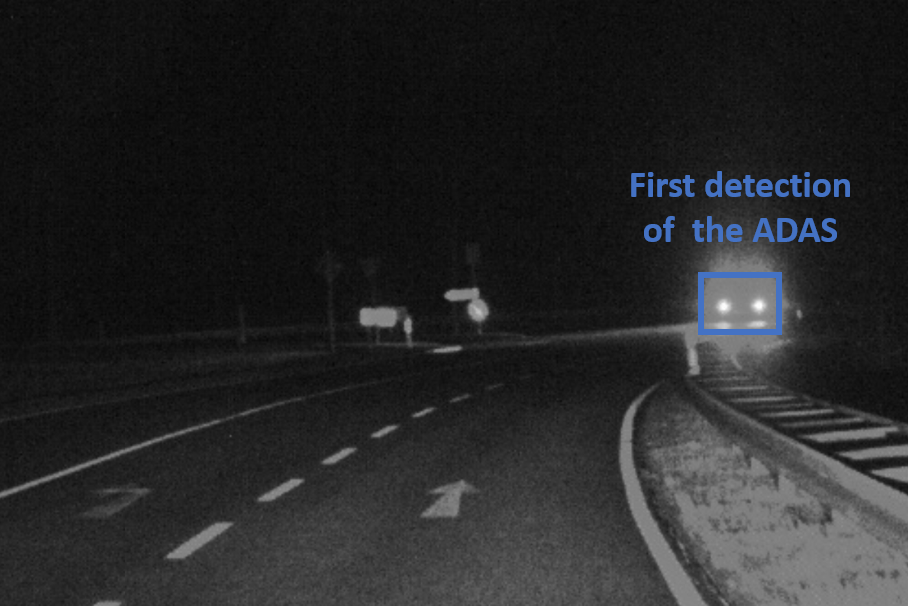}}
\par\end{centering}
\caption{Visual demonstration of the time benefit of the proposed system in
terms of a provident vehicle detection by the glare-free high beam
functionality on a recorded scene during test drives. The left images
always show the full image, and the right images always show a cropped
version of the full image. The upper part of all the images is cut
off so that the figure fits on one page.\label{fig:Visual-demonstration}}
\end{figure*}
\figref{Visual-demonstration} shows an example scene of the test
drives on rural roads at night. This scene illustrates very well the
accuracy and time benefit of the proposed system in a real environment
in terms of a provident vehicle detection compared to the in-production
computer vision system. In \figref{Visual-demonstration-start}, the
first light artifact of the oncoming vehicle can be seen. After 0.5\,s,
the first detection is made by the proposed system based on a single
frame, as shown in \figref{Visual-demonstration-single-frame}. Then,
after 2.6\,s, the tracker has validated the object and output's it
correctly to the glare-free high beam module, see \figref{Visual-demonstration-tracker}.
Based on the result of the tracker, the end of the road is dimmed
proactively (a black gap can be seen in the white box) to avoid blinding
the oncoming driver. The in-production system detects the oncoming
vehicle after 3.8\,s when it is fully visible and after a significant
latency, see \figref{Visual-demonstration-adas}. Therefore, the in-production
system would have caused a short glare for the oncoming driver. In
this scene, there is a total time benefit of 1.2\,s of the proposed
system.

This experiment provides a useful visualization interface of the detection
results in the real world. It also shows that despite the localization
uncertainties of the proposed detector and distance estimator, the
information can be used to realize a provident glare-free high beam
system. However, there are two points that need to be clarified: 
\begin{enumerate}
\item Why is the dimmed gap (see the white box in \figref{Visual-demonstration-tracker})
larger than the detected light reflection and tends to the left? 
\item Why does the tracker take ``so long'' after the first detection
to detect the oncoming vehicle? 
\end{enumerate}
First, the reason for the size of the dimmed area is due to the low
resolution of the headlights (see \subsecref{Experimental-platform}),
and, for safety reasons, detected objects are always increased by
a safety margin inside the glare-free high beam module. Moreover,
the left tendency might be caused by the inaccuracies of the distance
estimation. Second, the reason for the late detection of the tracker
(in this case, 2.1\,s after the first detection based on a single
frame) is because the vehicle disappears behind the trees several
times, which makes it difficult for the tracker to continuously track
the vehicle over multiple frames.

\section{Conclusion and outlook\label{sec:Conclusion-and-Outlook}}

Extending the work of \textcite{Oldenziel2020} and \textcite{Saralajew2021},
with this work, we presented a complete pipeline designed for automotive
use cases which is capable of providently detecting vehicles at night.
The system consists of a set of algorithms solving the tasks of detection,
three-dimensional localization, and tracking of both direct light
instances (e.\,g., headlights) and indirect light instances (e.\,g.,
light reflections on guardrails) caused by oncoming vehicles. The
evaluation shows that this detection pipeline can detect oncoming
vehicles almost 1.6\,s earlier than conventional vehicle detection
systems at night, which can be considered a significant amount of
time for automotive use cases. Also, by deploying the pipeline in
a test car for the use case of providently controlling the glare-free
high beam system for oncoming vehicles, the applicability of the proposed
detection pipeline is demonstrated not only under laboratory conditions
but also in real scenarios and in real time. 

Currently, for further use cases (e.\,g., trajectory planning, automatic
emergency braking), the system might still lack the necessary precision
in three-dimensional localization of the light reflections. Therefore,
future work should focus on evaluating new distance estimation methods
by extending the currently applied ground plane assumption to a more
precise representation of the environmental geometry ahead. Additionally,
in order to exploit the full potential of provident vehicle detection,
it is necessary to address the points circumvented in \subsecref{Provident-glare-free-high}.
One of the biggest challenges in this context might be to identify
the point where the oncoming vehicle might appear. In addition, future
work should also investigate the applicability of the provident detection
algorithm in urban scenarios (e.\,g., at intersections in cities).
If successful, this would be an important step on the way to computer
vision algorithms that resemble human perceptual capabilities.

\printbibliography

\end{document}